\documentclass[runningheads]{llncs}

\usepackage{eccv}

\usepackage{eccvabbrv}
\newcommand{\myparagraph}[1]{\paragraph{#1}}

\newcommand{\method}{\textsc{GimbalDiffusion}\xspace}

\usepackage{microtype}
\usepackage{multirow}
\usepackage{amssymb}
\usepackage[table]{xcolor} 
\usepackage{algpseudocode}
\usepackage{algorithm}

\usepackage{siunitx}
\usepackage{gensymb}
\usepackage{wrapfig}

\usepackage{xcolor}

\newif\ifeccv

\eccvtrue

\usepackage{graphicx}
\usepackage{booktabs}
\usepackage{enumitem}
\usepackage[accsupp]{axessibility}  % Improves PDF readability for those with disabilities.

\usepackage{hyperref}

\usepackage{orcidlink}

\begin{document}
\raggedbottom

% ---------------------------------------------------------------
% TODO REVIEW: Replace with your title
\title{GimbalDiffusion: Gravity-Aware Camera Control for Video Generation} 

% TODO REVIEW: If the paper title is too long for the running head, you can set
% an abbreviated paper title here. If not, comment out.
% \titlerunning{Abbreviated paper title}

% TODO FINAL: Replace with your author list. 
% Include the authors' OCRID for the camera-ready version, if at all possible.
\author{Frédéric Fortier-Chouinard$^1$, Yannick Hold-Geoffroy$^2$\orcidlink{0000-0002-1060-6941}, Valentin Deschaintre$^2$\orcidlink{0000-0002-6219-3747},
Matheus Gadelha$^2$\orcidlink{0000-0002-4971-7980}, Jean-François Lalonde$^1$\orcidlink{0000-0002-6583-2364}}
% \author{First Author\inst{1}\orcidlink{0000-1111-2222-3333} \and
%Second Author\inst{2,3}\orcidlink{1111-2222-3333-4444} \and
%Third Author\inst{3}\orcidlink{2222--3333-4444-5555}}

% TODO FINAL: Replace with an abbreviated list of authors.
\authorrunning{F.~Fortier-Chouinard et al.}
% First names are abbreviated in the running head.
% If there are more than two authors, 'et al.' is used.

\institute{Université Laval, Québec, Canada \and
Adobe, San Jose, USA\\
\email{frederic.fortier-chouinard.1@ulaval.ca}
\url{https://lvsn.github.io/GimbalDiffusion}
%\email{frederic.fortier-chouinard.1@ulaval.ca,\\\{holdgeof,deschain,gadelha\}@adobe.com,jflalonde@gel.ulaval.ca}
}
% TODO FINAL: Replace with your institution list.
% \institute{Princeton University, Princeton NJ 08544, USA \and
% Springer Heidelberg, Tiergartenstr.~17, 69121 Heidelberg, Germany
% \email{lncs@springer.com}\\
% \url{http://www.springer.com/gp/computer-science/lncs} \and
% ABC Institute, Rupert-Karls-University Heidelberg, Heidelberg, Germany\\
% \email{\{abc,lncs\}@uni-heidelberg.de}}

\maketitle
\newlength{\customwidth}

{
    \centering
    \includegraphics[clip,width=\textwidth]{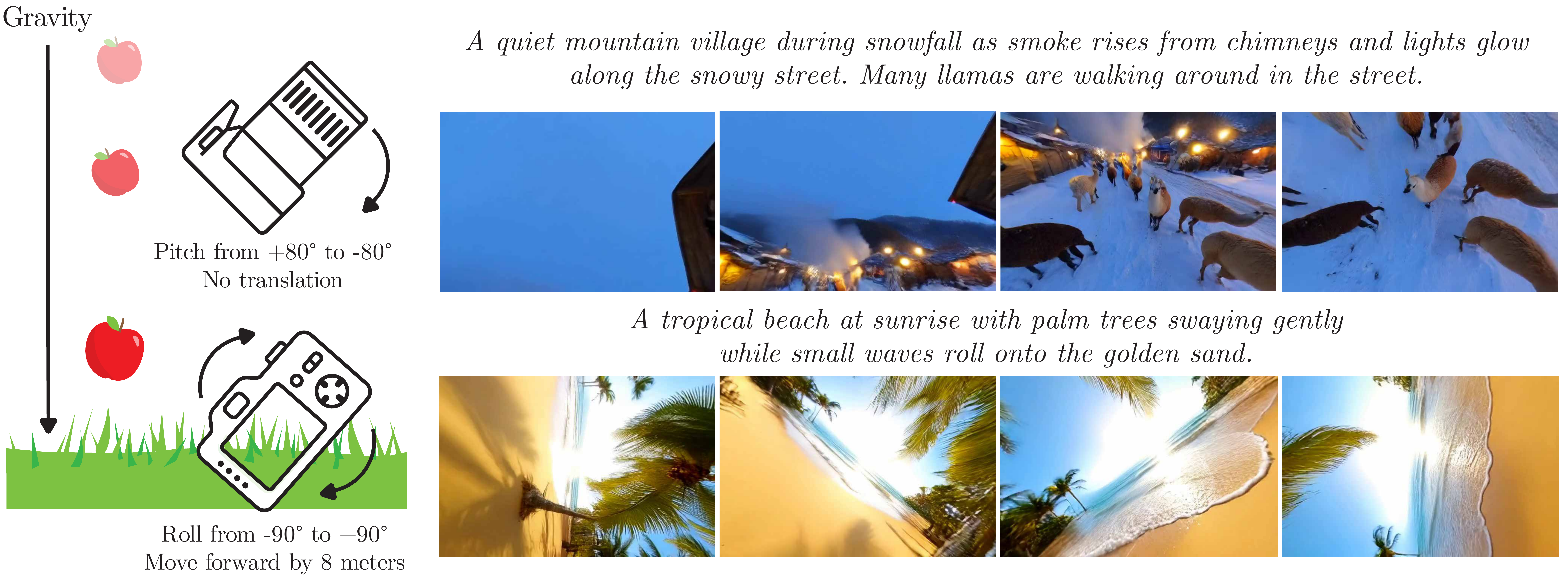}
    \vspace{-0.5cm}
    \renewcommand{\footnote}[1]{\protect\footnotemark{}}
    \captionof{figure}{
        We propose \method{}, a framework for absolute camera control in text-to-video generation. Our approach adapts foundational video generation models to accept absolute camera controls, conditioning the entire video on camera parameters expressed in a gravity-aligned global coordinate system. This enables the generation of videos with control over translation and extreme camera angles---such as very high pitch (top) or roll (bottom)---directly from text, something existing methods struggle to achieve. Here, the results were obtained with \method trained on WAN-2.2 5B with Plücker conditioning. Refer to the supplementary material for the full videos. 
    }
    \label{fig:Teaser}
}
\begin{abstract}

Recent progress in text-to-video generation has achieved remarkable realism, yet fine-grained control over camera motion and orientation remains elusive, especially with extreme trajectories (\eg, a 180\textdegree{} turnaround, or looking directly up or down). Existing approaches typically encode camera trajectories using relative or ambiguous representations, limiting precise geometric control and offering limited support for large rotations. 
We introduce \method{}, a framework that enables camera control grounded in physical-world coordinates, using gravity as a global reference. Instead of describing motion relative to previous frames, our method defines camera trajectories in an absolute coordinate system, allowing accurate, interpretable control over camera parameters. Using panoramic $360^\circ$ videos for training, we cover the full sphere of possible viewpoints, including combinations of extreme pitch and roll that are out-of-distribution of conventional video data. To improve camera control, we introduce null-pitch conditioning, a strategy that prevents the model from overriding camera specifications in the presence of conflicting prompt content (\eg, generating grass while the camera points toward the sky). 
Finally, we propose new benchmarks to evaluate gravity-aware camera-controlled video generation, assessing models' ability to generate extreme camera angles and quantify their input prompt entanglement. 

\ifeccv
\keywords{Video generation \and Diffusion models \and Camera control}
\fi

\end{abstract}

\section{Introduction}

Text-to-video generation is rapidly advancing, offering increasingly photorealistic results and new creative workflows. Yet, the ability to control generated video content---especially camera trajectory---remains limited, particularly for extreme camera rotations such as top-down shots, Dutch angles, or full 360\degree{} pans. Early approaches rely solely on textual input which does not provide fine-grained control, such as specifying geometric parameters for camera paths and angles.

Recent methods~\cite{he2024cameractrl,he2025cameractrl,bahmani2025vd3d,bahmani2025ac3d,bai2025recammaster,wan2025,li2025cameras,zhang2025unifiedcamerapositionalencoding} address this limitation by conditioning video models with explicit camera trajectories, frequently encoded as Plücker rays or relative camera positional encodings. However, these representations are inherently ambiguous and lack the precision of real camera systems. Typically, camera extrinsics used for Plücker conditioning are defined \emph{relative} to the first frame, making it impossible to explicitly control camera orientation with an absolute reference within the environment (\eg, with respect to gravity). Furthermore, they are trained on videos with limited camera movement and fail to generalize to extreme camera rotations. % As a result, these methods are limited to modest rotations and fail at extreme camera angles. 

\begin{figure*}
    \centering
    \scriptsize
    \setlength{\tabcolsep}{0.5pt}
    \setlength{\customwidth}{0.12\linewidth}
    \begin{tabular}{ccccccccc}
    & Forward & \multicolumn{2}{c}{Up} & \multicolumn{2}{c}{Back} & \multicolumn{2}{c}{Down} & Forward \\
     \rotatebox{90}{\scriptsize ~~Ours} &
       \includegraphics[width=\customwidth]{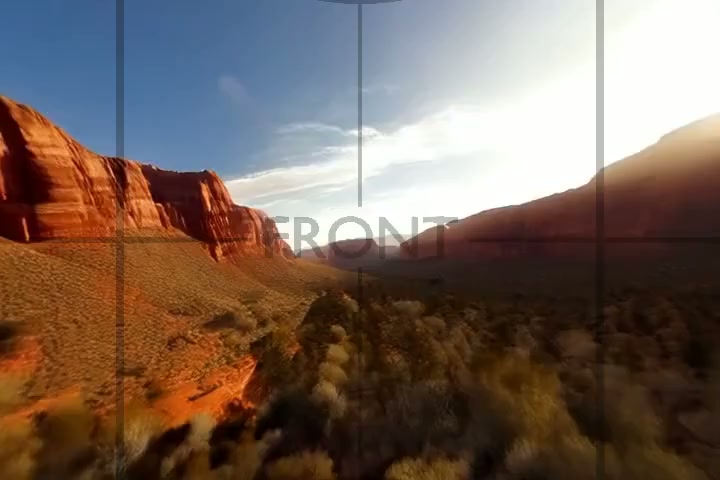}  & 
       \includegraphics[width=\customwidth]{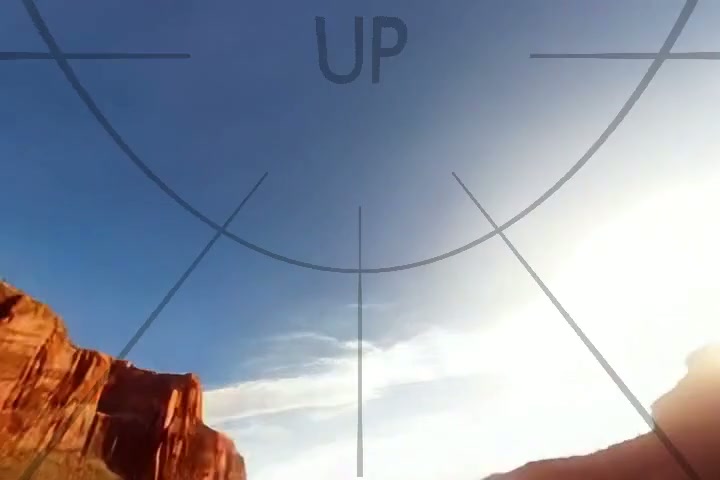} &\includegraphics[width=\customwidth]{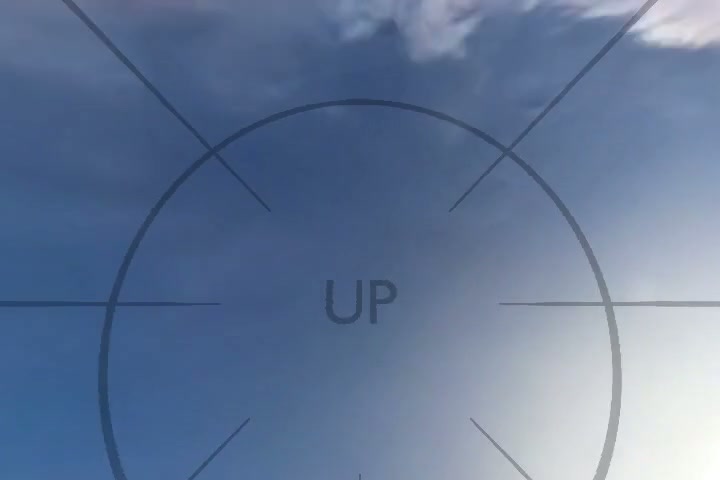} &
       \includegraphics[width=\customwidth]{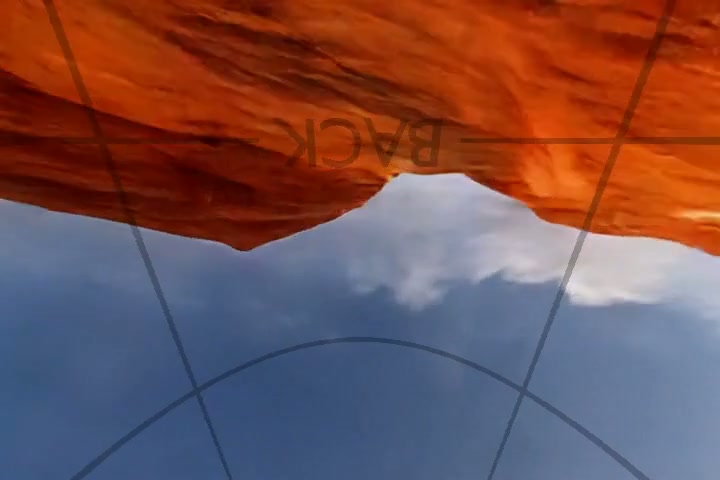}  & 
       \includegraphics[width=\customwidth]{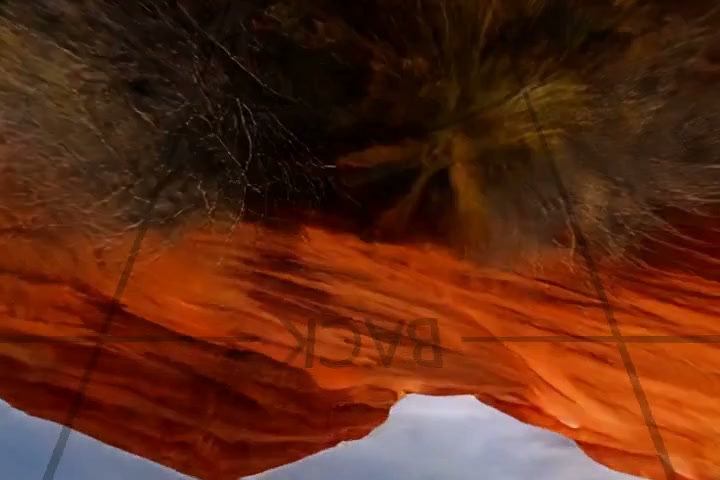}  & 
       \includegraphics[width=\customwidth]{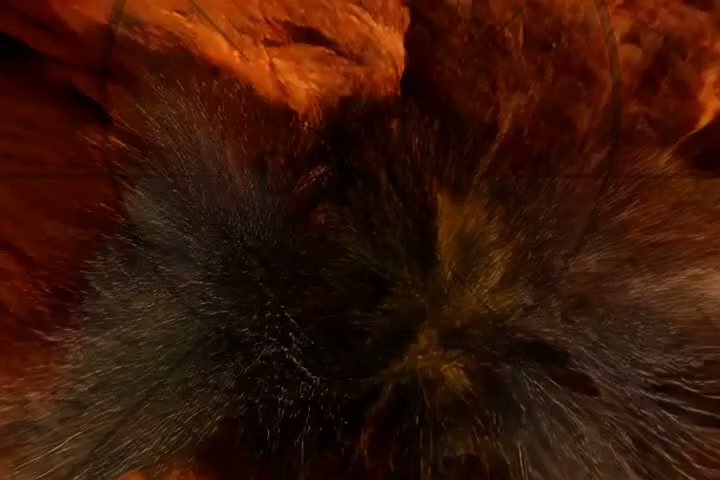}  & \includegraphics[width=\customwidth]{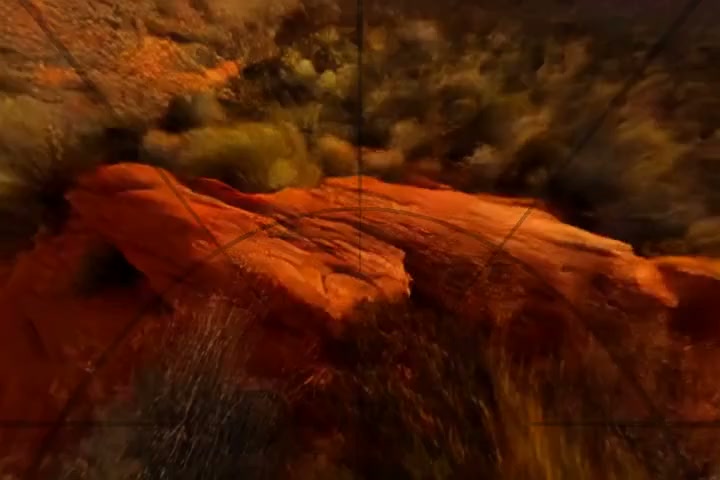} &
       \includegraphics[width=\customwidth]{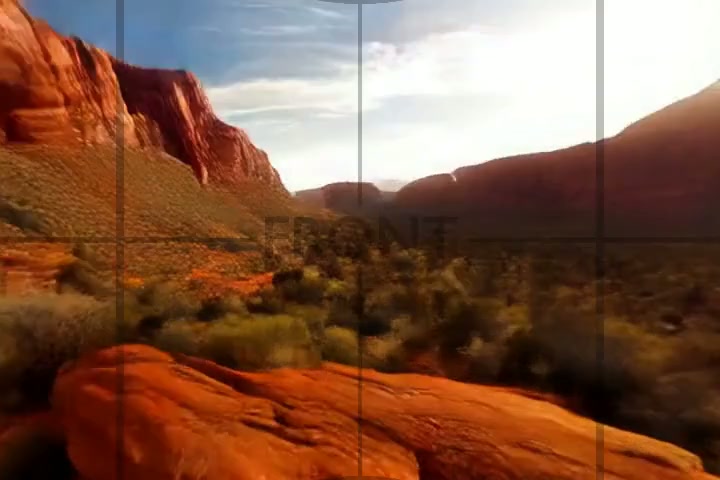} \\
     \rotatebox{90}{\scriptsize ~UCPE} &
       \includegraphics[width=\customwidth]{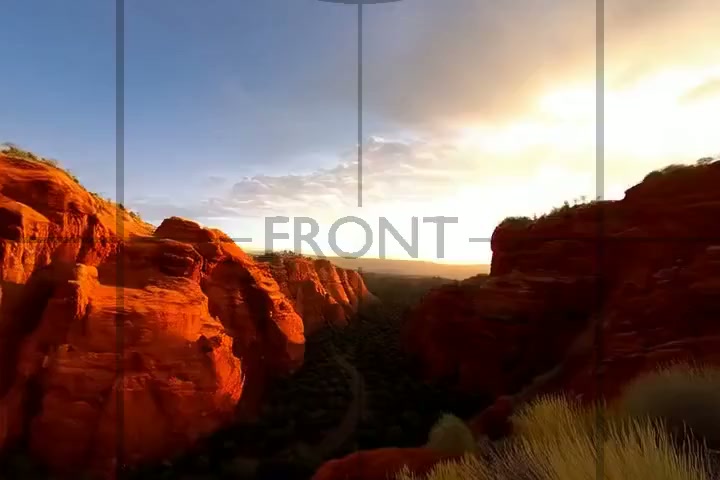}  & 
       \includegraphics[width=\customwidth]{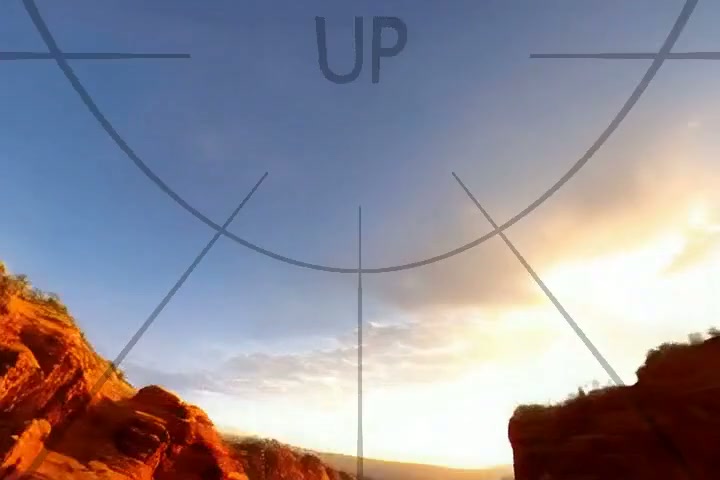} &\includegraphics[width=\customwidth]{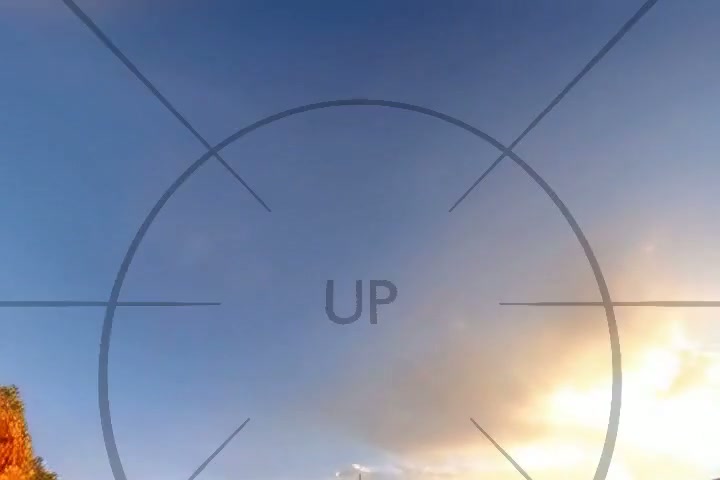} &
       \includegraphics[width=\customwidth]{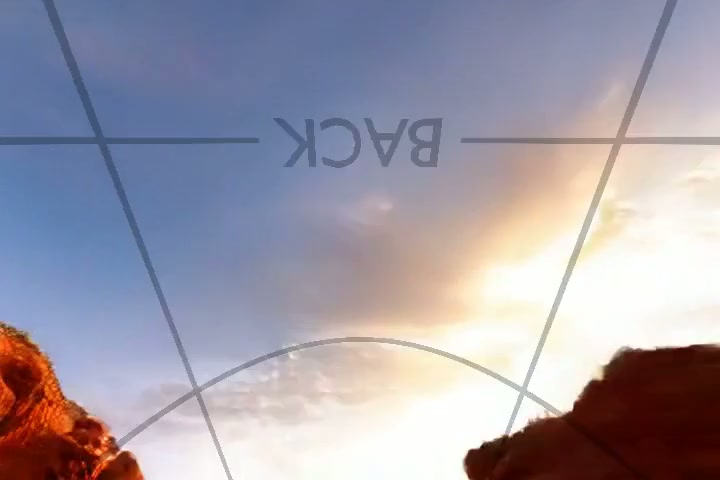}  & 
       \includegraphics[width=\customwidth]{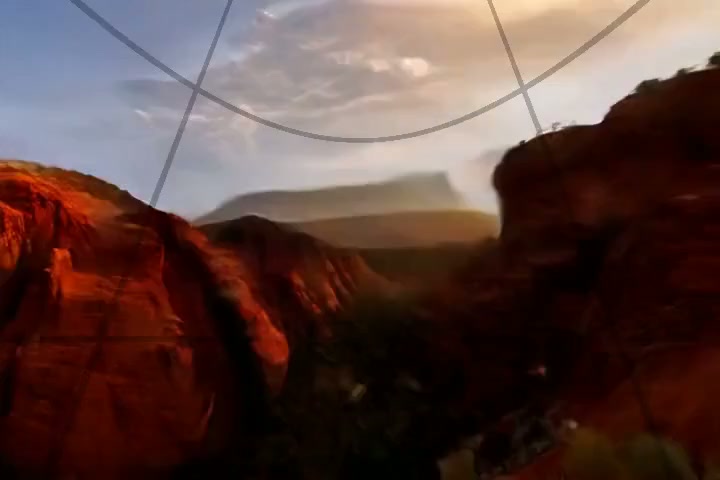}  & 
       \includegraphics[width=\customwidth]{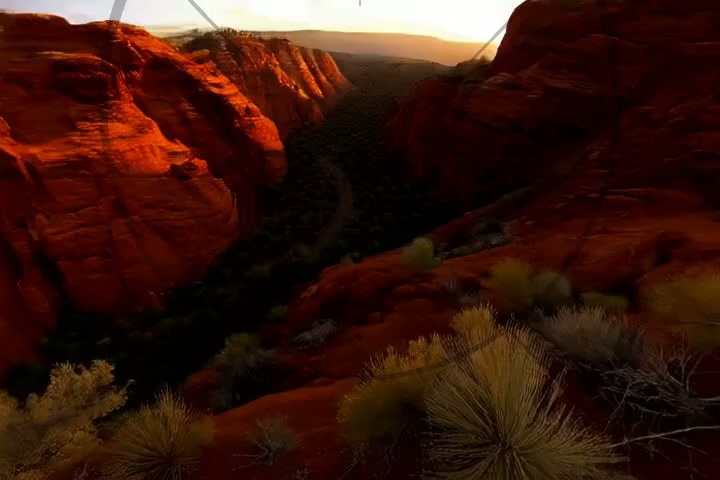}  & \includegraphics[width=\customwidth]{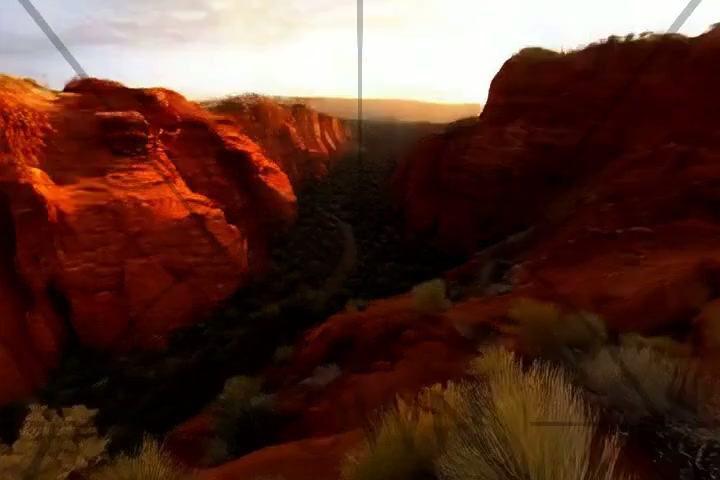} &
       \includegraphics[width=\customwidth]{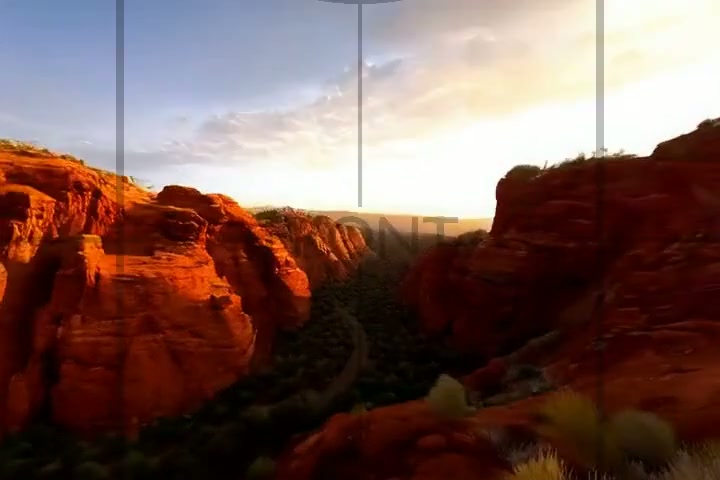} \\\rotatebox{90}{\scriptsize ~Gen3C} &
       \includegraphics[width=\customwidth]{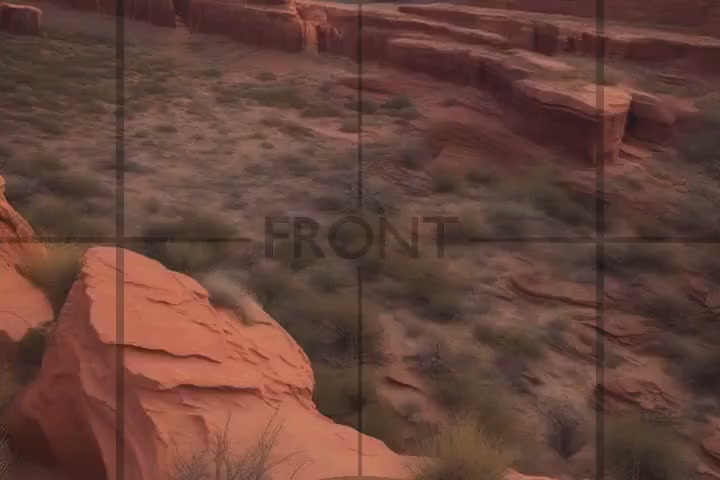}  & 
       \includegraphics[width=\customwidth]{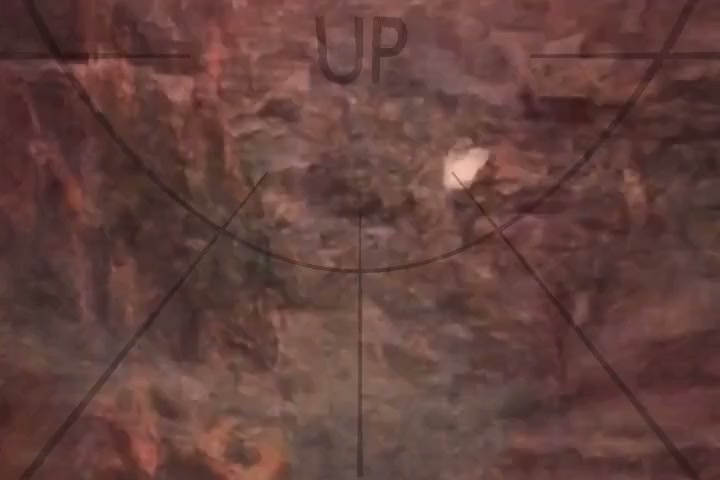} &\includegraphics[width=\customwidth]{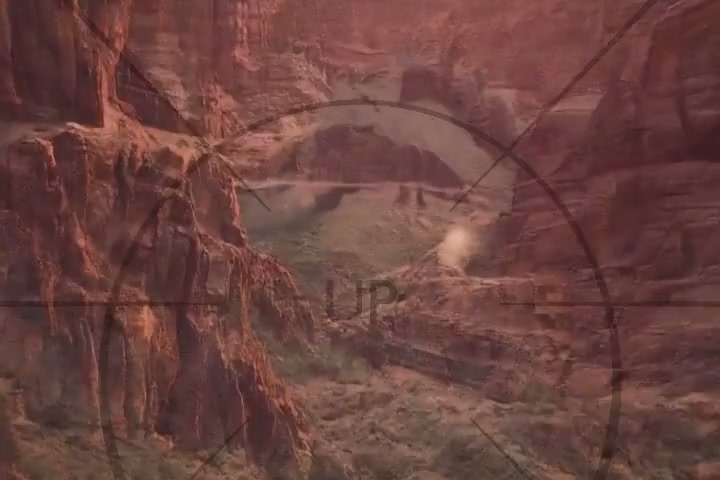} &
       \includegraphics[width=\customwidth]{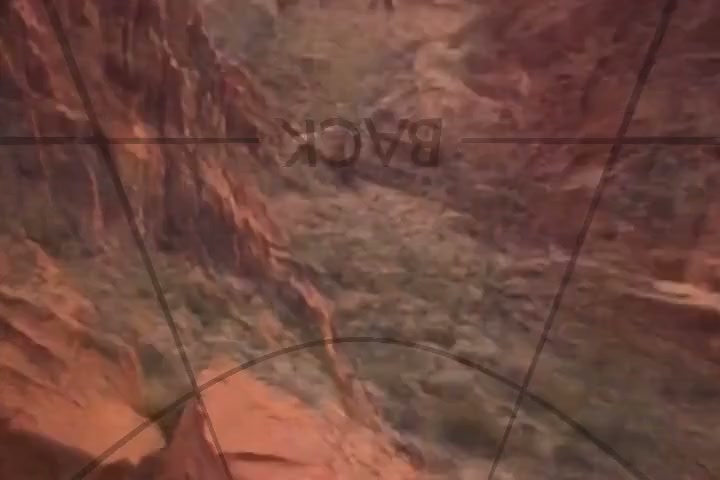}  & 
       \includegraphics[width=\customwidth]{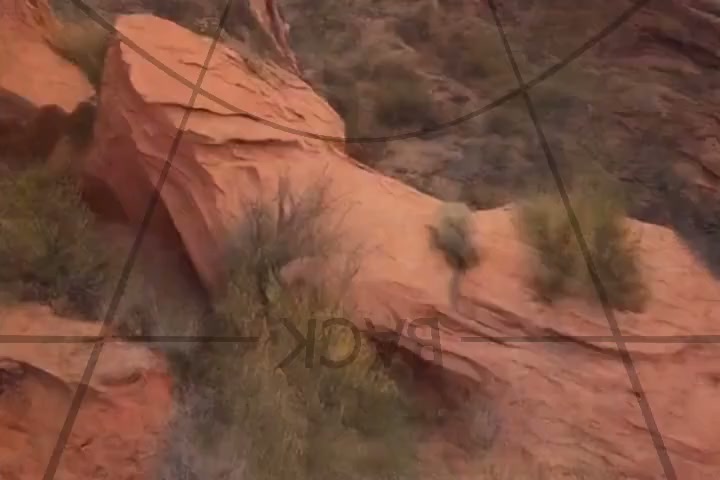}  & 
       \includegraphics[width=\customwidth]{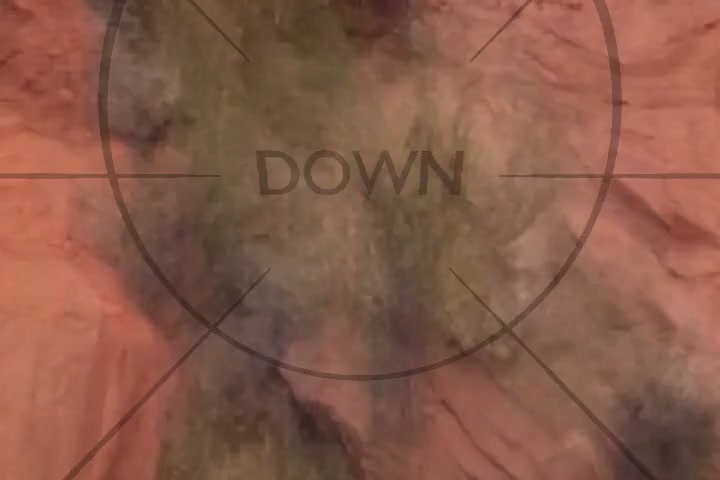}  & \includegraphics[width=\customwidth]{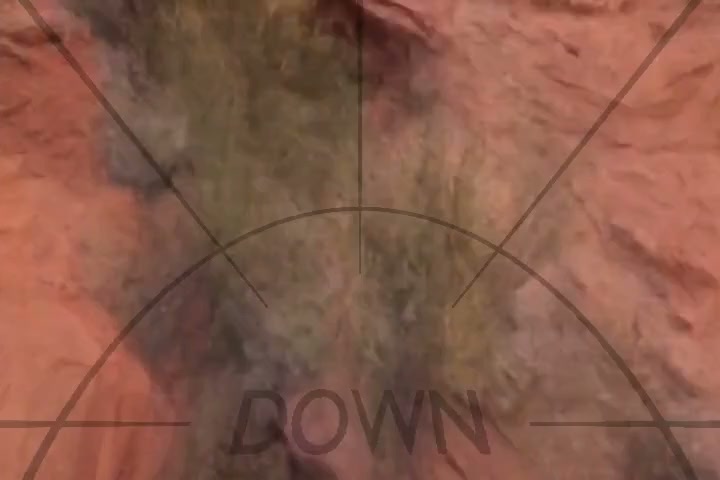} &
       \includegraphics[width=\customwidth]{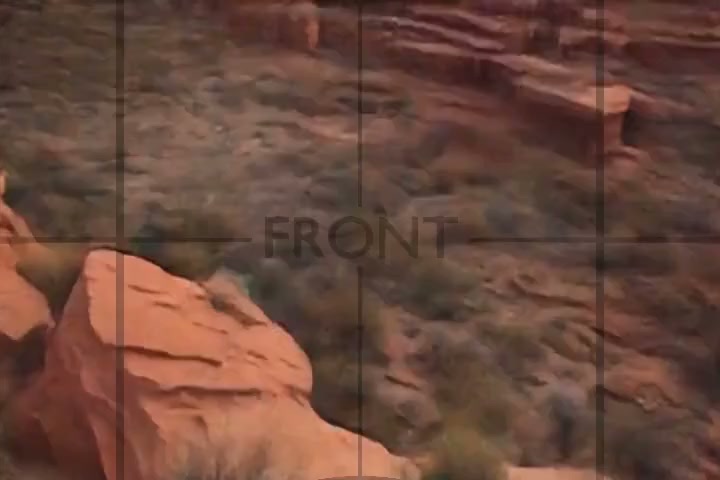} \\
       & \multicolumn{8}{p{0.98\linewidth}}{\scriptsize\centering``A dramatic canyon landscape with towering red rock cliffs carved by a winding river far below, warm sunset light illuminating layered rock formations, sparse desert vegetation [...]''}
    \end{tabular}
    \caption{Extreme rotation handling. Here, the camera starts forward and performs a motion akin to a ``backflip'': looking straight up, then backwards, down, and back to the starting position. Our method (top) precisely follows the desired camera trajectory while accurately reproducing the original content from the first frame. 
    Other methods, like UCPE~\cite{zhang2025unifiedcamerapositionalencoding} and GEN3C~\cite{ren2025gen3c} (initialized with PreciseCam~\cite{bernal2025precisecam}), fail at this task. Refer to the supp. for the full videos.}
    \label{fig:extreme-rotation}
\end{figure*}

In this work, we introduce \method{}, a novel gravity-centric approach to data annotation that overcomes this challenge and enables precise, absolute user control over generated camera trajectories, as illustrated in~\cref{fig:Teaser}. The name reflects the role of a gimbal---a device that stabilizes orientation in physical space---mirroring our use of gravity as a global reference to stabilize and disambiguate camera rotation. To resolve rotational ambiguity, we leverage geometrically calibrated 360\degree{} panoramic videos. By sampling camera crops over the sphere, we can remove camera bias in human-captured datasets, enabling the sampling of rare cinematic camera shots, such as barrel rolls (a ``twisting'' camera rotation), extreme low- and high-angle perspectives, or looping 360\degree{} rotations.
The effectiveness of this approach is demonstrated in \cref{fig:extreme-rotation}, where our method produces frames whose camera views more closely match the intended trajectories than those of previous methods. This enables models trained with our pipeline to learn a wide range of camera trajectories, far beyond the predominantly straight, forward-facing paths in conventional videos.

We further observe that text and camera angles are inherently entangled, and that care must be taken to ensure that appropriate control is achieved. For example, when asked to generate a video of a ``green field under a blue sky'' with a camera looking \emph{down}, a model must learn to render \emph{only} grass, even if other concepts are specified in the prompt. To this end, we propose a null-pitch conditioning approach in which captioning is performed on forward-facing crops rather than on the actual camera point of view. 

Finally, we propose two new evaluation benchmarks. The first evaluates video generation with extreme camera angles by leveraging the large-scale, high-quality dataset SpatialVID-HQ~\cite{wang2025spatialvidlargescalevideodataset} and rebalancing the camera pitch distribution to obtain diverse test samples. The second evaluates entanglement between prompt and camera angle by annotating crops of panoramas from the PolyHaven dataset~\cite{polyhavenHDRIs2025}. We will release these test datasets upon publication.

In summary, we make the following contributions:
\begin{itemize}[noitemsep]
    \item A pipeline for generating diverse camera trajectories from panoramic 360\degree{} video, providing exhaustive coverage of possible camera orientations;
    \item A strategy named \emph{null-pitch conditioning} for disentangling text from absolute camera angle, critical for extreme viewpoints where prompt content and camera orientation diverge most;
    \item Two evaluation benchmarks for: video generation with extreme camera angles, and assessing the entanglement between input prompt and camera orientation.
\end{itemize}

\section{Related Work}

\myparagraph{Video generation.}
Generative video modeling has evolved rapidly, from early GAN-based methods that produced short, low-resolution clips \cite{tulyakov2018mocogan,li2018video,Acharya2018} to large-scale diffusion- and flow-based architectures capable of generating temporally coherent, semantically consistent videos \cite{Weng2024}. Recent open-source models such as CogVideoX \cite{yang2024cogvideox} and WAN \cite{wan2025} push the frontier of video length, resolution, and controllability while proposing both text-to-video and image-to-video variants. 

Editing videos remains challenging: precise changes to content, viewpoint, or lighting are difficult once the frames have been synthesized. 
To address this, control strategies have emerged, introducing user-guided conditioning based on edge maps, depth, sketches, or explicit 3D geometry for structural control \cite{chen2023controlavideo,Yenphraphai2024ImageSculpting,chang2024magicpose,hu2023animateanyone}, and illumination-aware guidance for lighting control \cite{kocsis2024lightit,fortier2026spotlight,Magar_lightlab}. These mechanisms allow precise control over motion, composition, and scene appearance, enhancing the usability and compositional fidelity of generative video systems.

\myparagraph{Camera control methods.}
In static image generation, PreciseCam~\cite{bernal2025precisecam} (similarly, \cite{liao2026thinking}) provide text-to-image models with absolute control over camera pose by conditioning on roll, pitch, field of view, and lens distortion alongside textual prompts. For image-to-video generation, ViewCrafter~\cite{yu2024viewcrafter} and GEN3C~\cite{ren2025gen3c} generate novel views from a single input image, but cannot handle scene motion. Methods such as MotionCanvas~\cite{xing2025motioncanvas}, Diffusion-as-Shader~\cite{gu2025das}, and TrajectoryCrafter \cite{mark2025trajectorycrafter} maintain spatial coherence with point cloud conditioning from an initial reference image. While effective, these approaches depend on the reference frame's depth and pose estimation quality, limiting flexibility and generalization. 

In text-to-video synthesis, AC3D~\cite{bahmani2025ac3d} allows user-defined camera trajectories but conditions motion relative to the first frame only, leaving global rotational and translational ambiguities unresolved. Similarly, Li \emph{et al.}~\cite{li2025cameras} manipulate positional encodings to control viewpoint changes, yet still lack a consistent absolute reference such as gravity. Concurrent with our work, UCPE~\cite{zhang2025unifiedcamerapositionalencoding} extends positional encodings to encode a latitude up map. However, its data generation is complex and requires matching 360\degree{} videos with rotations extracted from a secondary dataset. 
It also cannot handle extreme camera angles (see \cref{fig:extreme-rotation}).

These works underscore the need for precise, absolute camera control that remains robust under extreme rotations. Our approach addresses this gap by aligning camera orientation with gravity and anchoring motion within a consistent world frame, allowing extreme camera rotations in text-to-video synthesis.

\myparagraph{Panorama video generation methods}
\cite{wang2024360dvd,xia2025panowan,zhang2025panflow} adapt video generation backbones for the task of generating 360\degree{} equirectangular videos. In addition to text, they can be conditioned on an input image~\cite{wang2024360dvd,zhang2025panflow}, optical flow~\cite{wang2024360dvd} and derotated flow~\cite{zhang2025panflow}. While gravity-aligned camera control can be achieved by cropping the desired field of view out of the generated panorama, this comes with several disadvantages: 1) most of the panorama pixels are discarded, resulting in severe resolution degradation; 2) the desired concepts described in the prompt can be cropped out and thus not appear in the video; and 3) current text-conditioned methods~\cite{xia2025panowan,wang2024360dvd} do not provide control over camera translation. We demonstrate these downsides in the supplementary material.

\myparagraph{Camera pose estimation.}

Estimating accurate camera poses from images is a long-standing problem in 3D vision. Classical structure-from-motion pipelines such as VisualSFM~\cite{wu2011visualsfm}, OpenMVG~\cite{moulon2016openmvg}, and COLMAP~\cite{Schonberger2016COLMAP} reconstruct camera intrinsics and extrinsics through feature matching, triangulation, and global bundle adjustment. More recent learning-based approaches jointly infer scene geometry and pose from data: VGGT~\cite{wang2025vggt} leverages transformer-based global geometry tokens for pose regression, while DUSt3R~\cite{wang2024dust3r} and MASt3R~\cite{leroy2024grounding} predict dense correspondence fields to recover metrically scaled 3D structure. MegaSAM~\cite{Jiang2024MegaSAM} extends these ideas to foundation scale, achieving state-of-the-art alignment accuracy. ViPE~\cite{huang2025vipe} extends these ideas to handle in-the-wild videos.

Despite their accuracy, these systems typically operate in arbitrary coordinate frames and lack physical grounding in gravity or a global world reference, limiting their absolute interpretability. To mitigate this, several works estimate the gravity direction from single images \cite{jin2023perspective,veicht2024geocalib,xian2019uprightnet}. However, acquiring datasets with gravity annotations often requires auxiliary sensors, such as accelerometers, or additional semantic information about the scene. Examples include EDINA~\cite{Do_2022_EgoSceneMSR}, Horizon Lines in the Wild~\cite{workman2016hlw}, and the KITTI Horizon dataset~\cite{kluger2020temporally}. ScanNet \cite{dai2017scannet} proposes a dataset of 3D scenes captured using an RGB-D sensor, but exhibits limited diversity, primarily focusing on small indoor environments. % such as offices, apartments, and bathrooms. 

To circumvent such constraints, we geometrically calibrate 360\textdegree{} videos from the PanoVid dataset~\cite{xia2025panowan}. 
By combining the up vector derived from the panoramic video with SfM-estimated intrinsics and extrinsics, we obtain a consistent absolute reference for conditioning video models on gravity-aware camera control.

\section{Method}

We aim to train a camera-conditioned model that supports extreme camera rotations, including full 360\degree{} trajectories, extreme pitch, and large roll, while maintaining precise control. To achieve this, we ground camera poses in an absolute, gravity-aligned coordinate system. We refer to this global gravity reference as an \emph{absolute} reference, in contrast to most methods that describe camera motion \emph{relative} to a previous frame. Because our absolute reference is aligned with gravity, pitch and roll are well defined, whereas yaw remains unconstrained. Throughout the paper, \emph{absolute rotation} therefore denotes pitch and roll expressed in this shared gravity-aligned reference. 

\begin{figure*}[t]
    \centering
    \includegraphics[width=0.99\linewidth]{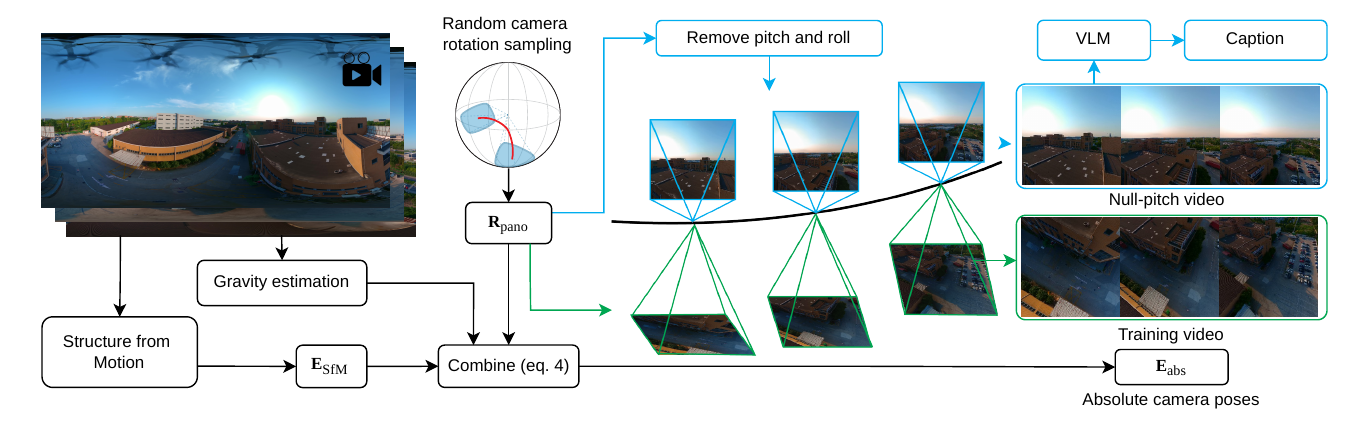}
    \caption{Training data pipeline. On the left, we start by estimating the camera poses and gravity up vectors from a dataset of 360\degree{} panoramic videos in equirectangular format (see \cref{sec:dataset}). We then randomly sample a camera rotation trajectory and camera intrinsics on the sphere, which are used for projecting the panoramic frames to perspective images. We integrate the two previous steps by combining the sampled rotation with the estimated camera poses (containing also translations). This process creates a dataset of perspective images (\textcolor[HTML]{00a651}{in green}) with their absolute camera poses. The main advantage of this approach is the ability to sample camera viewing directions across the entire sphere, in contrast to using natural human-captured videos, which are heavily biased towards null roll and pitch, and minimal rotation amplitude. We also project the panorama on the rotation trajectory with pitch and roll removed (\textcolor[HTML]{00aeef}{in blue}), yielding the null-pitch video from which we obtain a textual description (see \cref{sec:nullpitch}).}
    \label{fig:data_pipeline}

    % Notes about exporting from draw.io:
    % use settings here: https://github.com/jgraph/drawio/issues/4527#issuecomment-3743475160
  
    % draw io file: https://drive.google.com/file/d/1vghA5bboFir9r98WUMEJ8XBF7SYKmERe/view?usp=sharing
\end{figure*}

However, no large-scale video dataset provides the gravity annotations we need to train such a model. This section outlines our pipeline for obtaining such data, summarized in \cref{fig:data_pipeline}. We first extract camera pose annotations from a dataset of 360\degree{} panoramic videos (\cref{sec:dataset} and \cref{fig:data_pipeline}, left). Next, we sample a camera trajectory (\cref{sec:generating-camera-paths} and \cref{fig:data_pipeline}, center-bottom). Finally, we combine the two previous steps to produce our dataset. We repeat the same procedure for an additional version of each trajectory with pitch and roll set to 0\degree{} (\cref{sec:nullpitch}, \cref{fig:data_pipeline}, center-top). Before detailing the full pipeline, \cref{sec:representing_cameras} explains the standard representation to encode camera poses in neural-network-compatible buffers.

\subsection{Representing cameras}
\label{sec:representing_cameras}

    \method requires a gravity-aligned camera extrinsic matrix $\mathbf{E}_{\text{abs},f}\!=\![\mathbf{R}_{\text{abs},f} | \mathbf{t}_f]$ and intrinsic matrix $\mathbf{K}_f$ as input.
    Our method is agnostic to the specific camera encoding strategy: here, we showcase how it can be applied to training models conditioned on either UCPE~\cite{zhang2025unifiedcamerapositionalencoding} or Plücker rays (\eg, as in \cite{bahmani2025ac3d}) encodings.
    Unless otherwise noted, we use the UCPE encoding by default, which better maintains translation accuracy experimentally.

\myparagraph{Option 1: UCPE encoding.}
Recent work~\cite{miyato2024gta,kong2024eschernet,li2025cameras,zhang2025unifiedcamerapositionalencoding} has shown that camera intrinsics and extrinsics can be encoded for each token in a transformer, building upon the idea of Rotary Positional Embedding (RoPE)~\cite{roformer2024}.
UCPE~\cite{zhang2025unifiedcamerapositionalencoding} injects camera rays through lightweight trainable self-attention layers. 
For each token $k$, a block-diagonal matrix $\mathbf{D}_k^\text{UCPE}$ encodes its associated 3D ray transformation:
\begin{equation}
    \mathbf{D}_k^\text{UCPE} = \text{blkdiag}(\mathbf{D}_k^\text{Ray}, \mathbf{D}_k^\text{RoPE}) \,,
\end{equation}
where $\mathbf{D}_k^\text{Ray} = \mathbf{I}_\frac{d}{8} \otimes \mathbf{T}_k^{rw} \in \mathbb{R}^{\frac{d}{2} \times \frac{d}{2}}$ encodes the token's ray-to-world transformation as defined in~\cite{zhang2025unifiedcamerapositionalencoding}, and $\mathbf{D}_k^\text{RoPE}$ is the RoPE rotation matrix, determined by the token's position in time and 2D image space. The $\mathbf{D}_k^\text{UCPE}$ transformations are then applied in the self-attention following~\cite{miyato2024gta}:
\begin{equation}
    O = \mathbf{D} \odot \text{Attn}(\mathbf{D}^\top \odot Q, \mathbf{D}^{-1} \odot K, \mathbf{D}^{-1} \odot V) \,,
\end{equation}
where $\odot$ represents the token-wise matrix-vector product. Furthermore, unlike PRoPE~\cite{li2025cameras}, which is relative, UCPE encodes the latitude and up-vector through a linear layer and adds them to the self-attention input.

\myparagraph{Option 2: Plücker rays.}
Another common strategy for conditioning video generation models on viewpoint is to use Plücker rays~\cite{kant2024spad,he2024cameractrl,he2025cameractrl,bahmani2025vd3d,bahmani2025ac3d}. 
For a conditioning frame $f$, we can define a per-pixel map of Plücker rays as $\mathbf{p}_{f,u,v}\!=\!(\mathbf{t}_f\times \mathbf{d}_{f,u,v}', \mathbf{d}_{f,u,v}')$, where $\mathbf{t}_f$ denotes the camera origin in world coordinates, and $\mathbf{d}_{f,u,v}$ is the direction vector from the camera center toward the center of pixel $(u,v)$. Each pixel is thus represented as a 6D vector describing a ray in 3D space. The ray direction can be computed as $\mathbf{d}_{f,u,v}\!=\!\mathbf{R}_f\mathbf{K}_f^{-1}[u,v,1]^{T}$, with $\mathbf{d}_{f,u,v}'$ denoting the normalized direction vector $\mathbf{d}_{f,u,v}$.

\subsection{Creating varied camera paths from 360\degree{} videos}
\label{sec:dataset}

Most video datasets exhibit a significant bias toward forward-looking views with horizon level or near the center of the image, as shown in the supp. material.
However, such ``normal'' viewpoints limit storytelling expressivity by lacking cinematic shots, such as top-down, low-angle, or barrel-roll shots. Consequently, generative video models trained on such data struggle to produce these types of cinematic shots, as they lie outside the training distribution (see \cref{fig:extreme-rotation}). 
To address this limitation, we propose generating synthetic camera trajectories by sampling and rectifying 360\degree{} videos.

\paragraph{Camera poses from SfM.}

From a 360\degree{} video, we compute the \emph{relative} motion between frames by extracting six perspective crops (front, back, left, right, top, bottom), and processing the resulting video with a pose estimation method \cite{huang2025vipe}. This yields a set of camera poses $\mathbf{E}_{\text{SfM},f}$ for each frame $f$. Taking the first frame of the sequence $f=0$ as reference, we can then compute the relative motion 
\begin{align}
  \mathbf{E}_{\text{rel},f} &= \mathbf{E}_{\text{SfM},0}^{-1}\mathbf{E}_{\text{SfM},f} \; .
\end{align}

\paragraph{Gravity annotation.}
Pose estimation methods such as \cite{huang2025vipe} provide \emph{relative} estimates of camera rotations. To anchor the videos within a shared global reference frame, we align the relative camera rotations with gravity. Specifically, we use \cite{jin2023perspective} to estimate the camera roll and pitch angles at frame 0. Because it operates on perspective images, we first project the panoramic frame into eight perspective views evenly sampled about the yaw axis at fixed 90\degree{} field of view. We then estimate the camera up vectors from these views and rotate them back to the null yaw reference frame. We then average the up vectors across these views, yielding $\mathbf{\tilde{u}}_0$.
From this, we derive the absolute, gravity-aligned camera pose for each frame as
\begin{gather}
     \mathbf{t}_{\text{abs},f} =  \psi(\mathbf{\tilde{u}}_0)\mathbf{t}_{\text{rel},f} \nonumber \\
     \mathbf{R}_{\text{abs},f} =  \mathbf{R}_{\text{pano},f} \psi(\mathbf{\tilde{u}}_0)\mathbf{R}_{\text{rel},f} \\
     \mathbf{E}_{\text{abs},f} =  \varphi(\mathbf{R}_{\text{pano},0}) [\mathbf{R}_{\text{abs},f}|\mathbf{t}_{\text{abs},f} ] \;, \nonumber
\end{gather}
where $\varphi(\cdot)$ is a function that preserves only the roll and pitch angles (removes the yaw) of the first sampled camera pose, and $\psi(\cdot)$ aligns camera poses to the gravity up vector. After this adjustment, the transformed pose $\mathbf{E}_{\text{abs},f}$ has no rotation about the global up-vector axis. %This function $\varphi(\cdot)$ is obtained by computing the gravity direction in camera coordinates and inverting it to recover the corresponding up-vector (see the supplementary for the complete definition). 
Please refer to the supplementary material for the full definition of $\varphi(\cdot)$ and $\psi(\cdot)$.
This process establishes a gravity-aligned coordinate system consistent across all videos, enabling rotations to be defined within a common global frame. Consequently, each sequence is initialized with null yaw, reducing ambiguity and facilitating training with this representation. 

\begin{figure*}
    \centering
    \footnotesize 
    \setlength{\tabcolsep}{0.5pt}
    \setlength{\customwidth}{0.18\linewidth}
    \begin{tabular}{ccccc}
       \includegraphics[width=\customwidth]{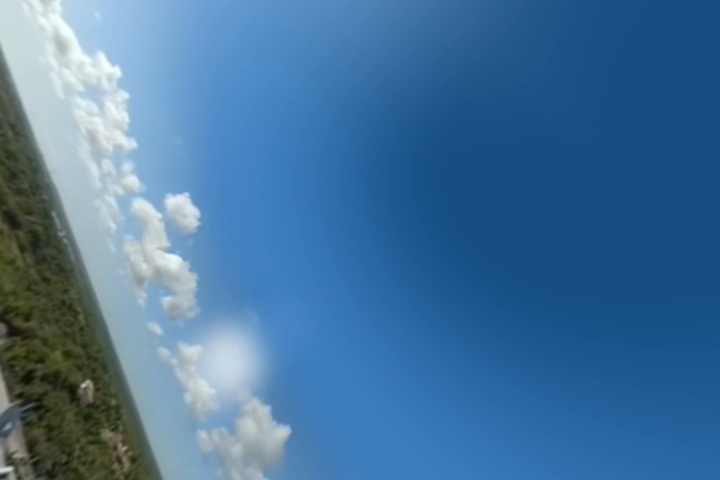}  & \includegraphics[width=\customwidth]{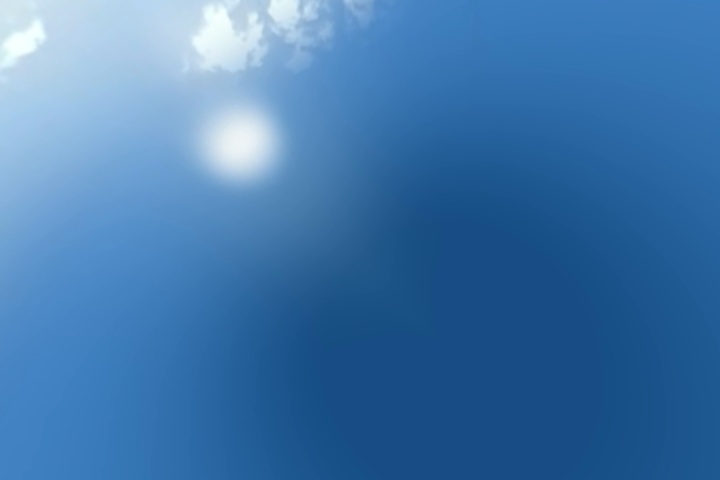} & \includegraphics[width=\customwidth]{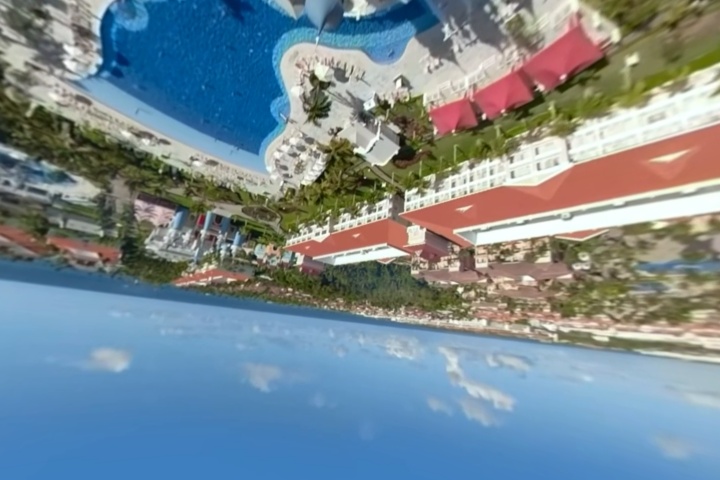} & \includegraphics[width=\customwidth]{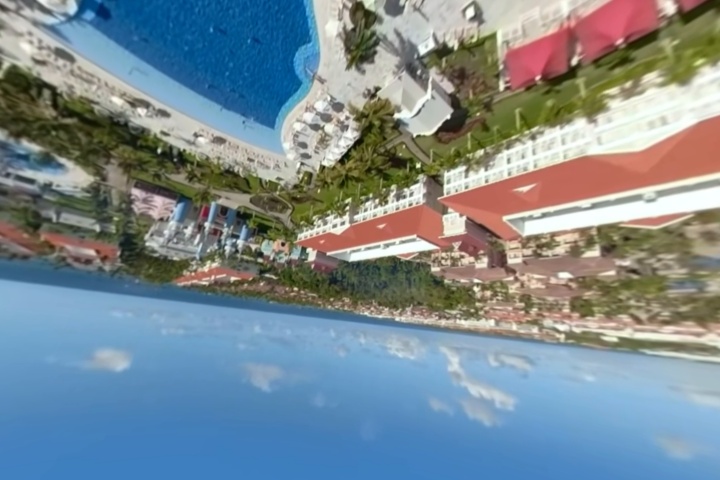} & \includegraphics[width=\customwidth]{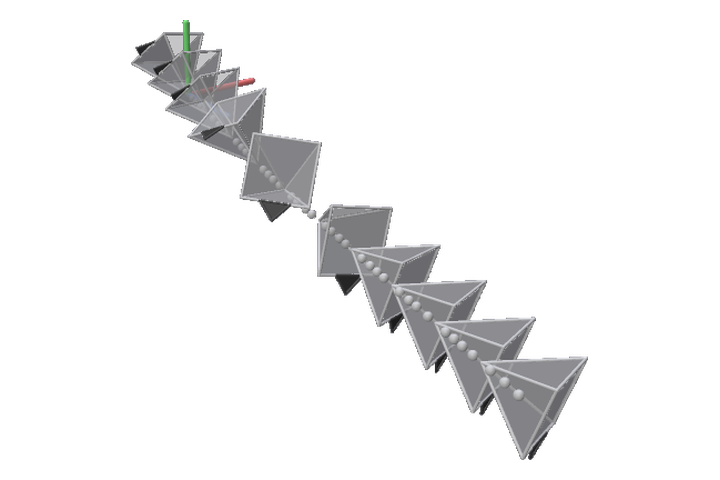} \\
       \multicolumn{4}{p{4\customwidth}}{\centering\tiny ``[...] an aerial view of a large resort complex featuring multiple buildings with red roofs [...]''} & \\
       \includegraphics[width=\customwidth]{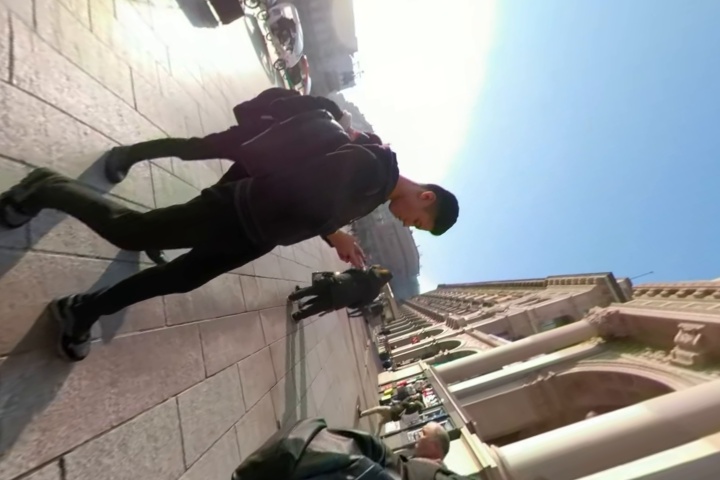}  & \includegraphics[width=\customwidth]{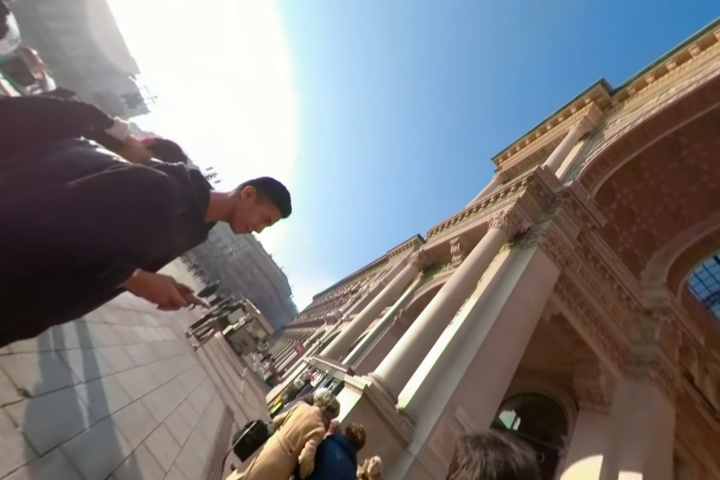} & \includegraphics[width=\customwidth]{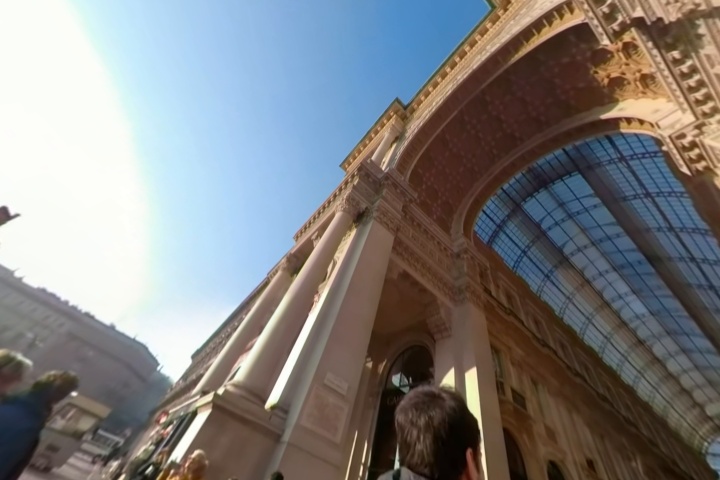} & \includegraphics[width=\customwidth]{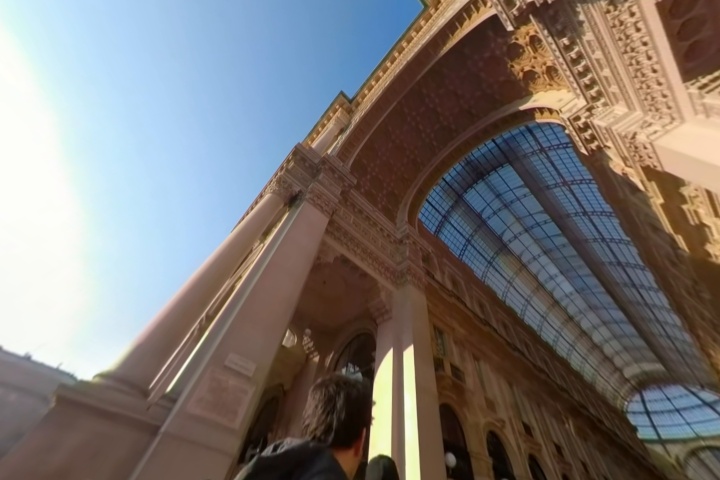} & \includegraphics[width=\customwidth]{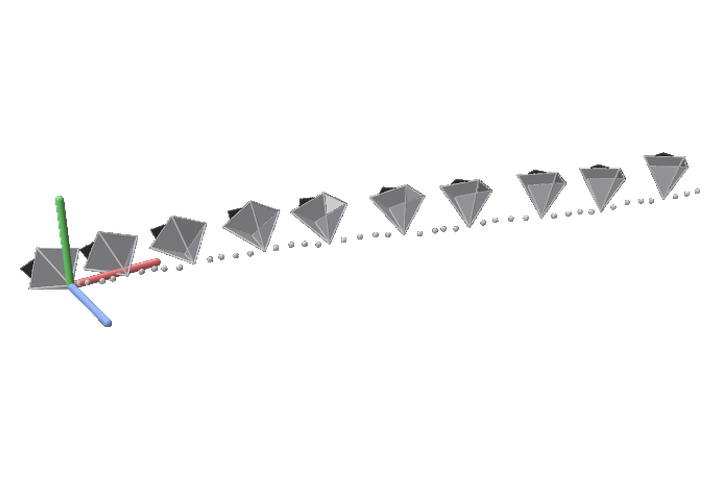} \\
       \multicolumn{4}{p{4\customwidth}}{\centering \tiny ``A young man in a black hoodie and pants walks along a [...] street, engrossed in his smartphone. [...]''} & \\
       \multicolumn{4}{c}{Videos and captions} & Camera poses \\
    \end{tabular}
    \caption{Training data samples from our data augmentation pipeline, capturing a highly  diverse set of rotation trajectories from 360\degree{} videos. Both the trajectory and the prompt are generated on-the-fly in our dataloader. See the supp. for full videos.}
    \label{fig:dataset_example}
\end{figure*}

\paragraph{Sampling random camera paths.}
\label{sec:generating-camera-paths}
We can generate several perspective videos from a 360\degree{} sequence by sampling random camera rotation paths. 
We outline our approach in \cref{algo:random_rot_sampling}. We first sample random pitch, roll, and yaw angles uniformly to obtain an initial viewpoint. Next, 1 to 3 additional random rotations are sampled and distributed across the sequence length. Those rotations are applied sequentially to produce smooth and varied motion paths. 

We further introduce variations in the camera field of view (FoV). For each sequence, we sample either 1, 2, or 3 keyframes at random timestamps, with an FoV ranging from 35\degree{} to 100\degree. The FoV values are then interpolated using cubic splines with randomized derivative boundary conditions. This approach provides much broader coverage of possible viewpoints than existing datasets (see supplementary). Notably, this sampling slightly biases viewpoints towards the poles, which we found to better support model stability and representation of more extreme viewpoints. Examples from our sampling are shown in \cref{fig:dataset_example}. 
\begin{algorithm}[ht]
\scriptsize
\caption{Random camera rotation path sampling}
\label{algo:random_rot_sampling}
\begin{algorithmic}
\State $\text{pitch}_0 \sim \text{Uniform}(-90\degree{}, 90\degree{})$ ; $\text{roll}_0 \sim \text{Uniform}(-90\degree{}, 90\degree{})$ ; $\text{yaw}_0 \sim \text{Uniform}(0\degree{}, 360\degree{})$
\State $\mathbf{R}_{\text{pano},f} \gets \left(\mathcal{R}_{\text{YXZ}}(\text{yaw}_0, \text{pitch}_0, \text{roll}_0)\right)_{f=0}^{F-1}$ \Comment{Initialize the camera orientation}

\State $N \sim \{0, 1, 2, 3\}$ \Comment{Random number of rotation axes}

\For{$i = 1$ to $N$}
    \State $d_t \sim \text{Uniform}(0, 1)$ \Comment{Random rotation duration}
    \State $t_s \sim \text{Uniform}(0, 1 - d_t)$ ; $t_e \gets t_s + d_t$ \Comment{Random start and end time}

    \State $\mathbf{a} \sim \mathbb{S}^2$ \Comment{Random unit axis of rotation}
    \State $\theta_{\max} \sim \text{Beta}(1.0, 5.0) \cdot 720^\circ \cdot d_t$ \Comment{Random angular displacement}
    
    \State Construct cubic spline $s(t)$ on $[t_s, t_e]$ with random boundary derivatives
    \vspace{0.2em}
    \For{$f = 0$ to $F-1$} \Comment{Loop over the $F$ frames}
        \State $\theta_f \gets \text{clamp}(s(f/F), 0, 1) \, \theta_{\max}$ \Comment{Compute the rotation angle}
        \State $\mathbf{R}_{\text{pano},f} \gets \exp([\theta_f \, \mathbf{a}]_\times) \times \mathbf{R}_{\text{pano},f}$ \Comment{Update the camera orientation}
    \EndFor
\EndFor

\State \Return $\mathbf{R}_{\text{pano}}$
\end{algorithmic}
\end{algorithm}

\subsection{Null-pitch conditioning}
\label{sec:nullpitch}

We observed that captioning videos using their actual \emph{rotated} crops causes the model to overemphasize the textual prompt when inferring the camera angles, rather than strictly adhering to the camera conditioning. This entanglement hinders the achievement of extreme camera angles, where visual content diverges most from the prompt. For instance, if a prompt describes ground features such as grass but the camera should face the sky, the model often still outputs a downward-looking view. To mitigate this entanglement, we caption each video using an upright, forward-looking image with a 90\degree{} field of view, thereby strengthening the model’s adherence to the camera conditioning. We refer to this strategy as ``null-pitch conditioning''. Specifically, we generate two sets of crops (see \cref{fig:data_pipeline}): one using the original randomly sampled camera trajectory, and another with pitch and roll set to 0\degree{} (with yaw unchanged). As illustrated in \cref{fig:data_pipeline}, a trajectory that consistently points towards the ground would otherwise yield captions describing only ground features. A model trained on such captions would then expect these elements to appear in its outputs, even when the camera conditioning specifies a view of the sky, leading to semantically incorrect generations. Training with null-pitch captions, which provide descriptions of the full scene, makes it easier for the model to handle extreme camera angles. During training, the randomly sampled perspective crops are used for the flow or diffusion loss, while the null-pitch crops are used for captioning. %We show that this training strategy improves camera controllability while maintaining prompt adherence, as shown in \cref{fig:entanglement_prompt_pitch}.

While this strategy helps the video model in handling extreme angles, we found it still struggled with undesirable elements present directly below the camera in real 360\degree{} training videos, such as tripods or the hand of a person holding a selfie stick.
Indeed, since these elements are never present in the training caption, the model learns to generate these objects as if they were normal scene content. To prevent this, we generate a third set of videos (not shown in \cref{fig:data_pipeline}), this time looking straight down (-90\degree{} pitch), and caption them with the VLM. We append the resulting ``look-down'' caption to the null-pitch caption 50\% of the time.
At inference time, this ``look-down'' prompt is used in the classifier-free guidance as a negative prompt to ensure that undesirable content is absent from the generated videos. We found this strategy to be most useful when fully finetuning a model, which tends to lead to overfitting on unwanted artifacts present in the training set. Please see the supplementary for additional details.

\section{Experiments}

This section presents the training data, details model training and losses, introduces our two novel evaluation benchmarks, presents baselines, and shows quantitative and qualitative experimental results on our benchmarks.

\subsection{Training data}

We use the ``YouTube videos'' subset of the PanoVid dataset~\cite{xia2025panowan}, which contains a diverse set of high-quality 360\degree{} videos, and filter out videos where the shortest side is less than 900 pixels. We obtain relative camera poses and per-frame up-vectors using the procedure from \cref{sec:generating-camera-paths}. We discard videos where pose estimation either failed or resulted in extreme mean acceleration values, typically indicating unreliable camera poses. Our filtering strategy reduces the number of training videos from 7797 to 4798 with high-quality camera poses (38\% discarded).

During training, we sample a random camera trajectory of 49 frames using \cref{algo:random_rot_sampling} at a resolution of $720 \times 480$ (or a similar resolution compatible with the backbone, see supp.), effectively acting as an infinite data augmentation strategy. 
We use InternVL-3-2B~\cite{zhu2025internvl3} for captioning, providing it with 6 evenly spaced frames across the video timeline. This is also done during training since new videos from different trajectories are sampled each iteration. Prompt lengths are further sampled between short, medium and long.

\subsection{Models and baselines}

We compare \method against recent diffusion- and flow-based methods for video generation with camera control. Since our method is invariant to the choice of architecture and can be applied to different camera conditioning mechanisms (\eg, UCPE and Plücker), we also train our method for each video generation baseline---except GEN3C, since it does not support dynamic scenes. We provide additional training and architecture details in the supplementary.

\myparagraph{AC3D~\cite{bahmani2025ac3d}.}  A camera encoder is trained to encode Plücker rays and inject them into the video DiT backbone. A parallel copy of 8 layers of the backbone is also trained, in a ControlNet fashion. The ControlNet is only applied for the first 40\% of denoising steps. Since AC3D is trained on camera poses relative to the first frame, we provide the absolute pitch and roll information through text (see supp. for the prompt template). We also test AC3D by feeding it with absolute camera poses (``+abs. Plücker''), even though this is outside its training distribution.

We compare the original checkpoint from~\cite{bahmani2025ac3d} trained on CogVideoX-2B by training \method in a similar fashion along with our contributions. In our case, since the model is trained on absolute Plücker rays, we do not concatenate the absolute camera information through text.

\myparagraph{PreciseCam+WAN~\cite{wan2025}.} A straightforward approach to camera-controlled video generation is to first use a camera-conditioned image generator to synthesize the initial frame, and then apply an image-to-video model to produce the full sequence. This baseline follows that strategy: PreciseCam~\cite{bernal2025precisecam} generates the first frame conditioned on the absolute pitch, roll, and field of view, and then Wan2.1-Fun-V1.1-1.3B-Control-Camera~\cite{wan2025} prolongs that first frame to a video. The video model is a fully finetuned checkpoint of WAN-2.1 1.3B, with an additional Plücker camera encoder and first frame input.

For \method, we train two variants, one using the same WAN-2.1 1.3B backbone and the other WAN-2.2 5B. For the former, we keep the original camera encoder architecture, whereas for the latter, we use a smaller camera encoder to reduce the weight count.

\myparagraph{PreciseCam+GEN3C~\cite{ren2025gen3c}.} GEN3C~\cite{ren2025gen3c} takes as input a 3D point cloud from a single image or an input video which can then be rendered from novel camera views.  We reuse the original implementation which predicts the depth using MoGE~\cite{wang2025moge} and feed it the first frame from PreciseCam~\cite{bernal2025precisecam}. We then project the point cloud along the relative camera poses and feed them to GEN3C. Note however that GEN3C can only generate novel views of static scenes, and therefore cannot produce dynamic video content (\eg, walking people). 

\myparagraph{UCPE~\cite{zhang2025unifiedcamerapositionalencoding}} is a concurrent method that also generates videos in a gravity-aligned reference frame, using WAN-2.1 1.3B as the backbone. It injects both absolute and relative camera orientation through additional self-attention layers. Unlike ours, UCPE requires a limited auxiliary dataset to simulate rotation diversity within panoramas, resulting in a less diverse dataset than our proposed approach.

For \method, we train the same backbone and conditioning strategy combined with our contributions.

\subsection{Video camera control benchmark}
There exists no standard evaluation dataset containing paired text prompts, camera poses, and reference videos that provide extreme camera angles. Most methods~\cite{bahmani2025ac3d, he2025cameractrl} use test splits from video datasets like RealEstate10K~\cite{RealEstate10K}, which have very limited diversity in terms of pitch and roll angles. 
Please refer to the supplementary material for a diversity analysis and quantitative results on RealEstate10K.

We propose a new evaluation benchmark, featuring diverse, high-quality videos, distributed with near-uniform pitch and roll coverage. We start with 371K video clips from the ``HQ'' subset of the SpatialVID~\cite{wang2025spatialvidlargescalevideodataset} dataset, which still exhibit a bias toward forward-looking orientations. We estimate the average pitch from the videos using Perspective Fields~\cite{jin2023perspective} and organize them into 10-degree bins spanning -85\degree{} to +85\degree{}. We manually filter out videos where the pitch estimation clearly failed until we obtain a subset of 138 videos with a uniform pitch distribution. For the roll, we found that even among the 371K videos, nearly all have near-zero roll. We thus artificially add a roll angle by randomly sampling a roll trajectory between -40\degree{} and 40\degree{} and applying it by warping the video. We name the resulting dataset \emph{SpatialVID-extreme}. By having a diverse set of pitches and rolls, we ensure a diverse set of text prompts and reference videos for FID and FVD computations. 
Please refer to the supplementary material for an evaluation of the diversity of our benchmark dataset. 

For the camera trajectories fed to the models, we annotate translations using ViPE~\cite{huang2025vipe} and normalize them to have a unit-length maximum translation magnitude (since some methods are not trained on metric-scale data), but discard the rotation. We instead generate a random rotation by sampling roll, pitch, and yaw trajectories (see the supp.\ for details). This strategy allows for a broader range of rotations and also better tests the methods' adherence to the camera trajectory. Indeed, we receive a normal environment description as input prompt, but for extreme camera angles, such as looking straight up, adhering to the camera conditioning likely requires generating views of the sky.

\subsection{Metrics}

To evaluate camera rotation controls, we first measure absolute orientation accuracy using two metrics: angular pitch error (``PitchErr''), and gravity error (``GravityErr''), defined as the angle between the ground-truth up vector and that of the generated video, in camera space. The pitch angle and up vector are obtained from each frame using Perspective Fields~\cite{jin2023perspective}. Roll error is not directly computed, as estimating it near the poles (looking straight up or down) is unreliable; however, roll deviations are implicitly reflected in the gravity error. 

To evaluate the relative rotation error (``RotErr''), we run VGGT~\cite{wang2025vggt} on the generated frames and compute the average relative angular error. We reuse the same definition of rotation error as \cite{he2024cameractrl}. 
To evaluate camera motion accuracy, we compute a translation error (``TransErr'') by comparing the estimated and ground-truth trajectories; both trajectories are normalized prior to computing the error to account for scale differences. 
The CLIP score \cite{hessel2021clipscore} is computed to measure the alignment between the generated video frames and the reference text description.
Finally, we employ the Fréchet Inception Distance (FID) and Fréchet Video Distance (FVD) to assess the visual quality and realism of the generated frames and sequences, respectively.

\begin{table*}[t]
    
\newcommand{\graymidrule}{%
  \arrayrulecolor{gray!40}\midrule
  \arrayrulecolor{black}%
  \addlinespace[-0.1em]%
}

\centering
\caption{Quantitative results on our SpatialVID-extreme evaluation dataset.
Our method, applied on different backbones and camera encodings, outperforms all baselines on absolute camera orientation metrics (PitchErr and GravityErr), 
and nearly all dynamic video generation methods on relative rotation error (RotErr), and is competitive on aesthetic quality (FID, FVD). Furthermore, removing the null-pitch conditioning drastically degrades absolute orientation metrics. PreciseCam + GEN3C is \textcolor{gray}{grayed out} to emphasize it cannot generate dynamic content.
}
\tiny

\label{tab:quant_results}
\begin{tabular}{lcccccccc}
\toprule
Method & Dynamic & PitchErr$_\downarrow$ & GravityErr$_\downarrow$ & RotErr$_\downarrow$ & TransErr$_\downarrow$ & CLIP$_\uparrow$ & FID$_\downarrow$ & FVD$_\downarrow$ \\
\midrule
\multicolumn{9}{l}{\textcolor{gray}{\emph{Cosmos 7B + projected point cloud encoder + full finetuning}}} \\
\textcolor{gray}{PreciseCam~\cite{bernal2025precisecam} + GEN3C~\cite{ren2025gen3c}}  & \textcolor{gray}{$\times$} & \textcolor{gray}{26.65} & \textcolor{gray}{30.80} & \textcolor{gray}{10.99} & \textcolor{gray}{0.40} & \textcolor{gray}{20.2} & \textcolor{gray}{101.2} & \textcolor{gray}{844.1} \\
\graymidrule
\multicolumn{9}{l}{\emph{CogVideoX-2B + Plücker camera encoder + ControlNet}} \\
AC3D~\cite{bahmani2025ac3d} + cam. text. & $\checkmark$ & 39.09 & 44.13 & 29.27 & \textbf{0.62} & 23.1 & 113.0 & 984.1 \\
AC3D~\cite{bahmani2025ac3d} + cam. text. + abs. Plücker & $\checkmark$ & 38.49 & 42.93 & 34.85 & 0.64 & \textbf{23.3} & 117.6 & 1022.6 \\
Ours & $\checkmark$ & \textbf{19.67} & \textbf{23.31} & \textbf{18.55} & 0.67 & 21.2 & \textbf{110.8} & \textbf{905.8} \\
\graymidrule
\multicolumn{9}{l}{\emph{WAN 2.1 1.3B + Plücker camera encoder + full finetuning}} \\
PreciseCam~\cite{bernal2025precisecam}+WAN-I2V-CC~\cite{wan2025} & $\checkmark$ & 29.16 & 33.48 & \textbf{24.54} & \textbf{0.62} & \textbf{22.3} & \textbf{109.2} & \textbf{776.3} \\
Ours & $\checkmark$ & \textbf{10.50} & \textbf{13.05} & 25.84 & 0.68 & 20.5 & 114.7 & 1041.6 \\
\graymidrule
\multicolumn{9}{l}{\emph{WAN 2.2 5B + Plücker camera encoder + full finetuning}} \\
Ours w/o null-pitch cond. & $\checkmark$ & 14.39 & 16.36 & 17.26 & \textbf{0.74} & \textbf{22.0} & \textbf{110.6} & \textbf{895.2} \\
Ours & $\checkmark$ & \textbf{8.36} & \textbf{10.54} & \textbf{15.05} & \textbf{0.74} & 20.6 & 116.0 & 1014.4 \\
\graymidrule
\multicolumn{9}{l}{\emph{WAN 2.1 1.3B + UCPE encoding + UCPE module training}} \\
UCPE~\cite{zhang2025unifiedcamerapositionalencoding} & $\checkmark$ & 16.25 & 18.88 & 18.38 & \textbf{0.44} & 22.5 & \textbf{110.7} & 957.9 \\
Ours w/o null-pitch cond. & $\checkmark$ & 19.13 & 22.07 & \textbf{11.68} & 0.48 & \textbf{22.9} & 114.3 & 1056.7 \\
Ours w/o absolute rotations & $\checkmark$ & 51.16 & 56.92 & 15.15 & 0.53 & 21.9 & 115.7 & \textbf{952.4} \\
Ours & $\checkmark$ & \textbf{13.95} & \textbf{16.95} & 12.26 & 0.46 & 21.5 & 114.0 & 997.9 \\
\bottomrule
\end{tabular}

\end{table*}

\subsection{Results} 

We present the quantitative results of our evaluation in \cref{tab:quant_results}. Overall, our method outperforms all dynamic video generation baselines on absolute camera orientation metrics (``PitchErr'', ``GravityErr'') and nearly all on relative rotation (``RotErr''). Since GEN3C can only synthesize novel viewpoints without scene motion, we include its results here for reference only. Our method achieves FID scores similar to those of other baselines and FVD scores on par with UCPE and AC3D. Baselines using PreciseCam as a starting point improve the FVD score, at the cost of much worse camera orientation metrics. We observe that our method results in a slight drop in CLIP score, see the discussion on prompt-camera entanglement in \cref{sec:prompt-camera-eval} below.

Representative qualitative results are illustrated in \cref{fig:qualitative} (see the supplementary for more). We observe that all baselines struggle with extreme angles (\eg, the ``down'' starting point in the bottom-right example). Here, they attempt to generate visual content in accordance with the prompt, but this contradicts the desired camera orientation. 
Another example is shown in the top-left example, where the ceiling is only correctly generated with our method when the camera is pointed up, whereas UCPE, PreciseCam+GEN3C and AC3D+cam. text. generate the wall.

\begin{figure*}[t]
    \centering
    \tiny
    \setlength{\tabcolsep}{2pt}
    \resizebox{\textwidth}{!}{%
    \begin{tabular}{cccc@{\hspace{4mm}}cccc}
    \textbf{UCPE} & \textbf{PreciseCam + GEN3C} & \textbf{AC3D + cam. text.} & \textbf{Ours} & \textbf{UCPE} & \textbf{PreciseCam + GEN3C} & \textbf{AC3D + cam. text.} & \textbf{Ours} \\
    \hline
    \includegraphics[width=0.27\textwidth]{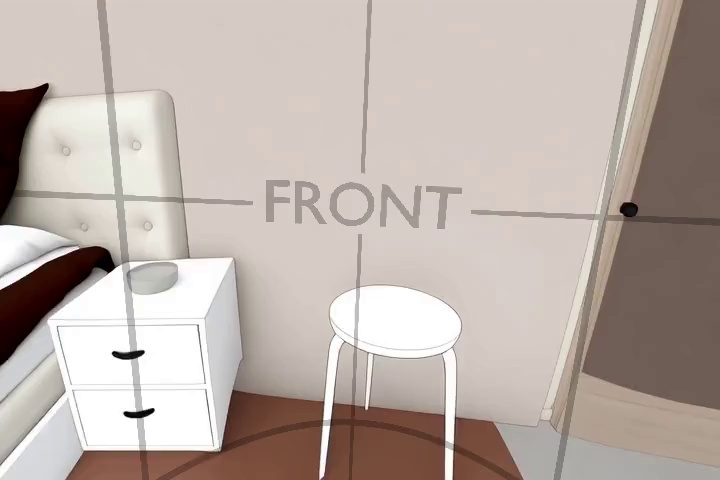} &
    \includegraphics[width=0.27\textwidth]{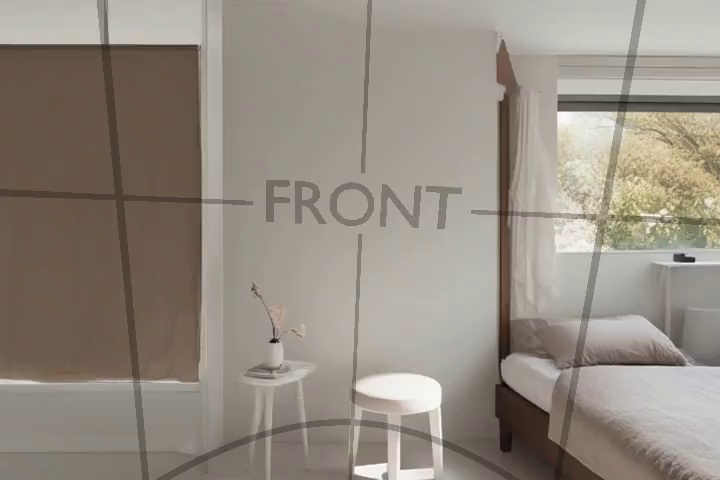} &
    \includegraphics[width=0.27\textwidth]{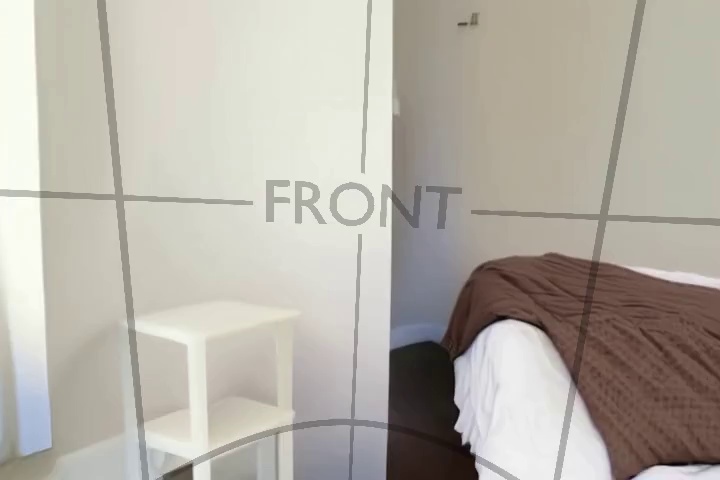} &
    \includegraphics[width=0.27\textwidth]{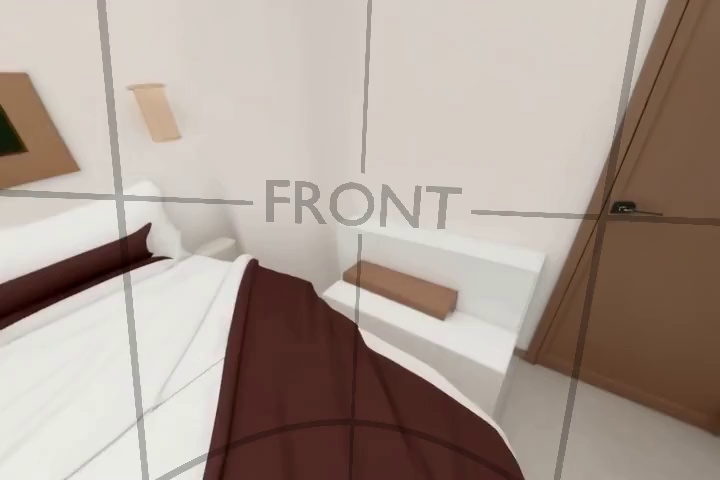} &
    \includegraphics[width=0.27\textwidth]{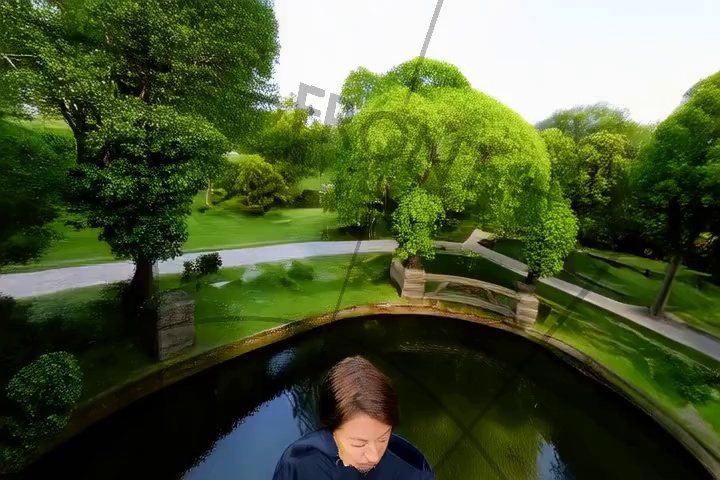} &
    \includegraphics[width=0.27\textwidth]{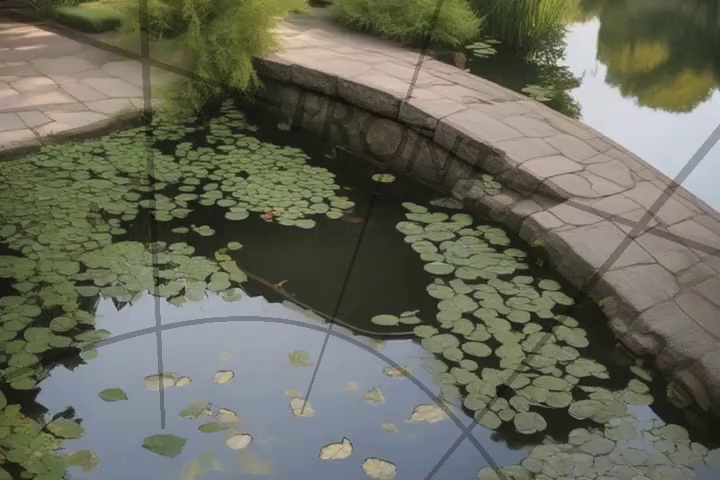} &
    \includegraphics[width=0.27\textwidth]{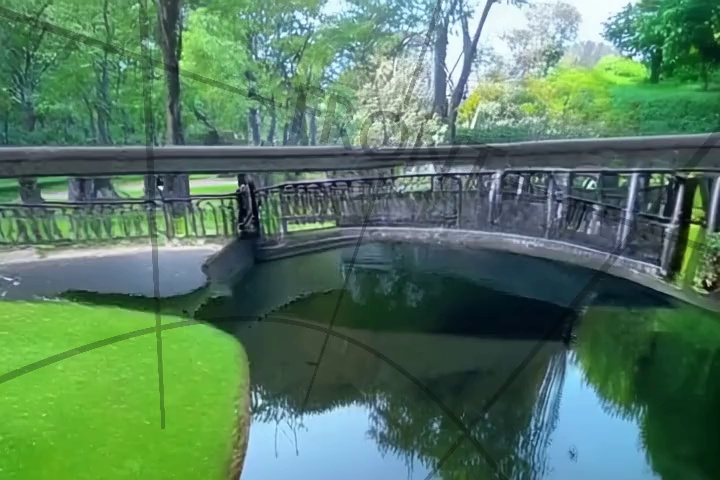} &
    \includegraphics[width=0.27\textwidth]{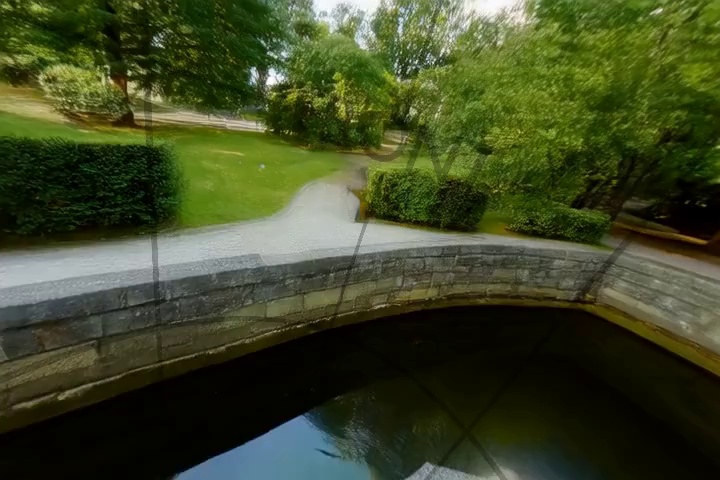} \\
    \includegraphics[width=0.27\textwidth]{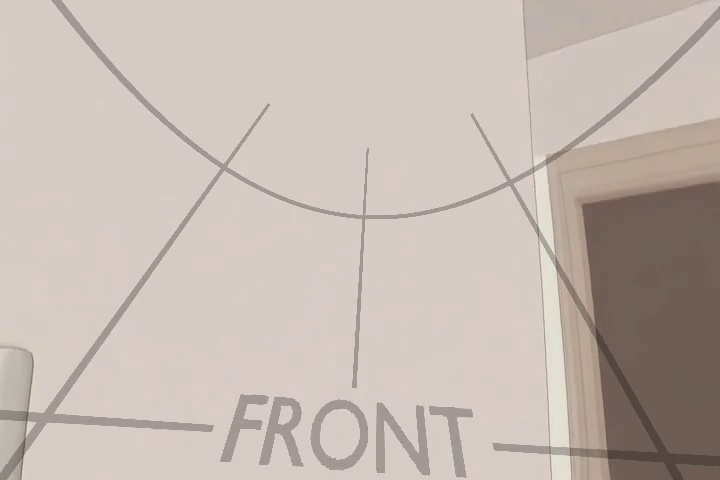} &
    \includegraphics[width=0.27\textwidth]{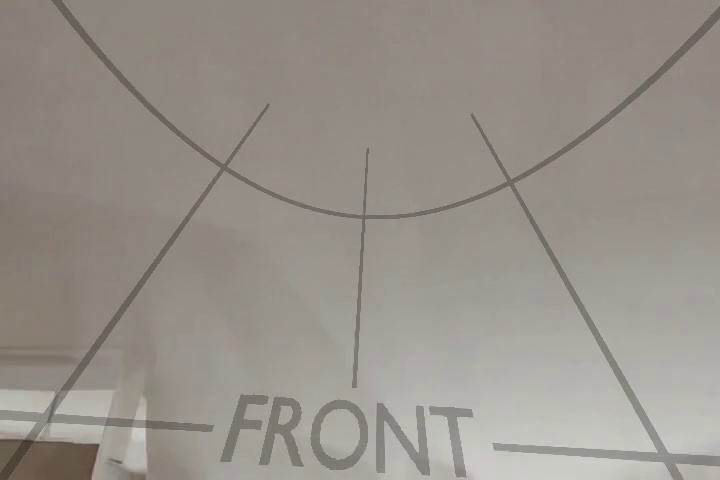} &
    \includegraphics[width=0.27\textwidth]{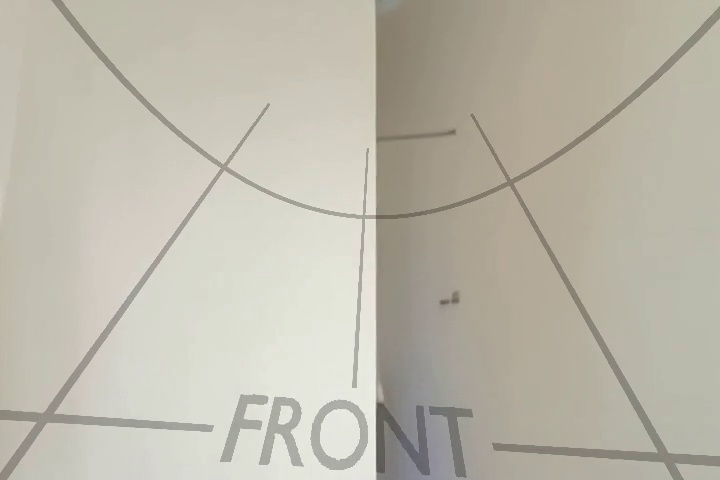} &
    \includegraphics[width=0.27\textwidth]{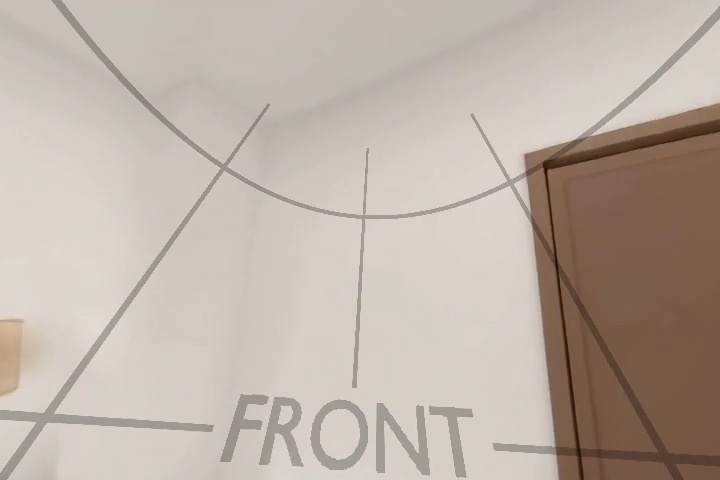} &
    \includegraphics[width=0.27\textwidth]{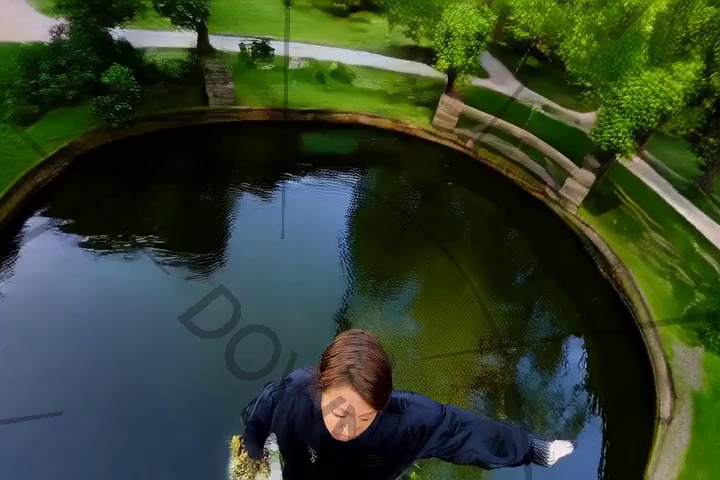} &
    \includegraphics[width=0.27\textwidth]{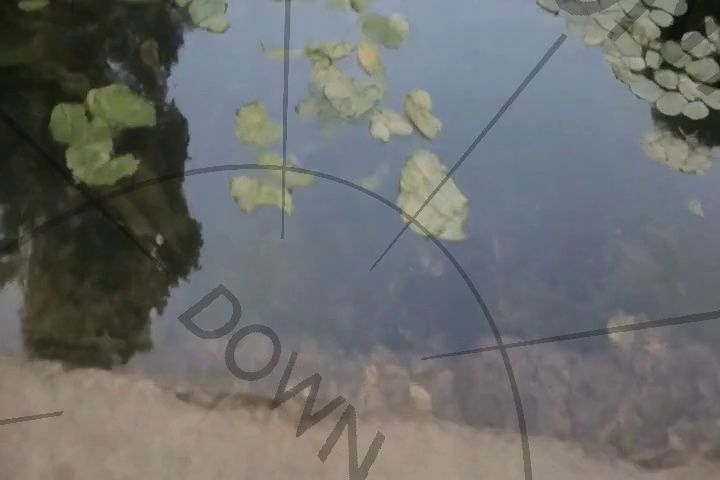} &
    \includegraphics[width=0.27\textwidth]{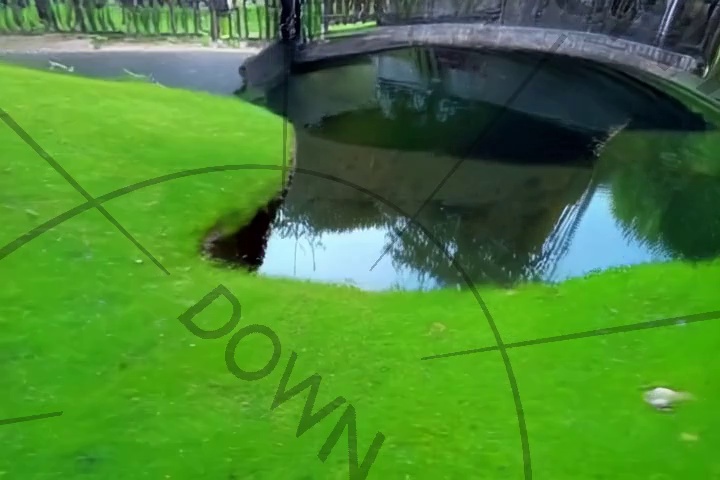} &
    \includegraphics[width=0.27\textwidth]{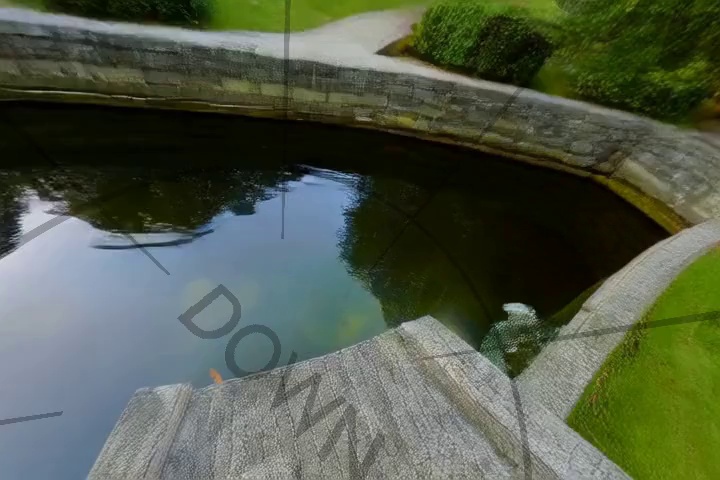} \\
    \includegraphics[width=0.27\textwidth]{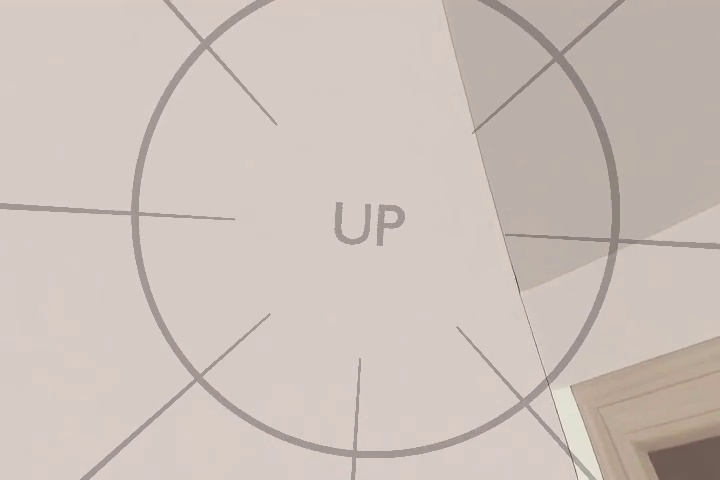} &
    \includegraphics[width=0.27\textwidth]{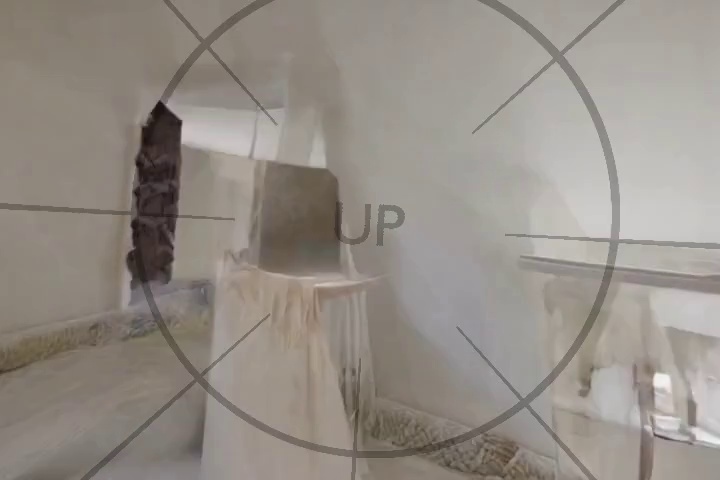} &
    \includegraphics[width=0.27\textwidth]{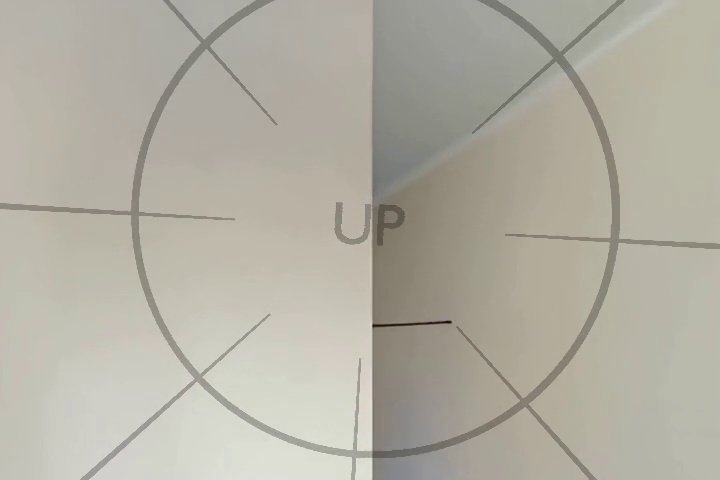} &
    \includegraphics[width=0.27\textwidth]{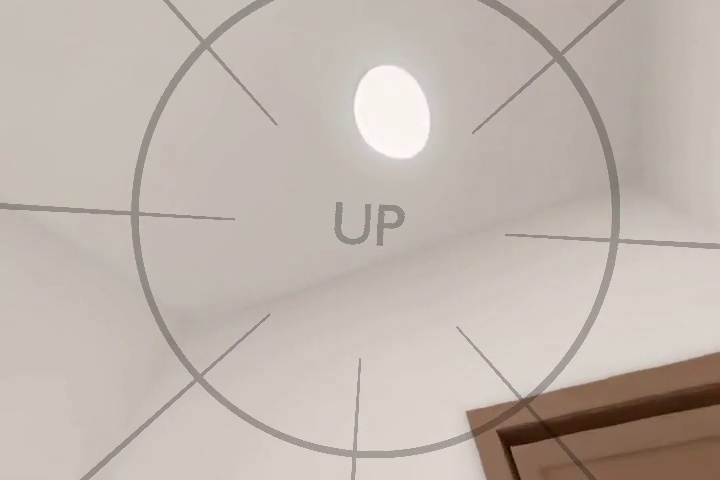} &
    \includegraphics[width=0.27\textwidth]{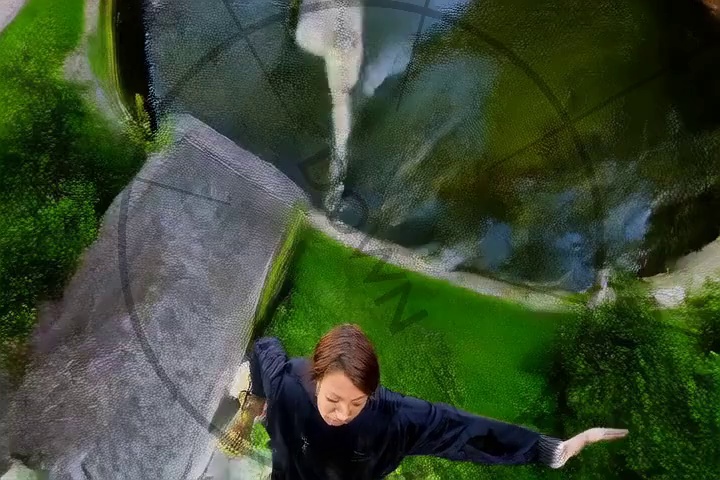} &
    \includegraphics[width=0.27\textwidth]{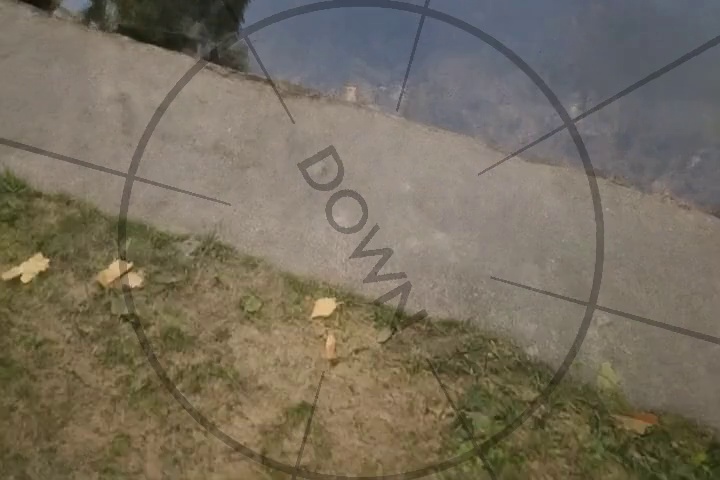} &
    \includegraphics[width=0.27\textwidth]{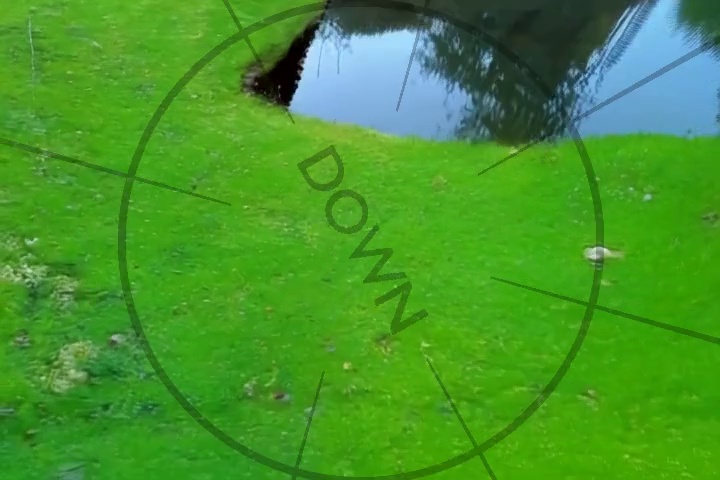} &
    \includegraphics[width=0.27\textwidth]{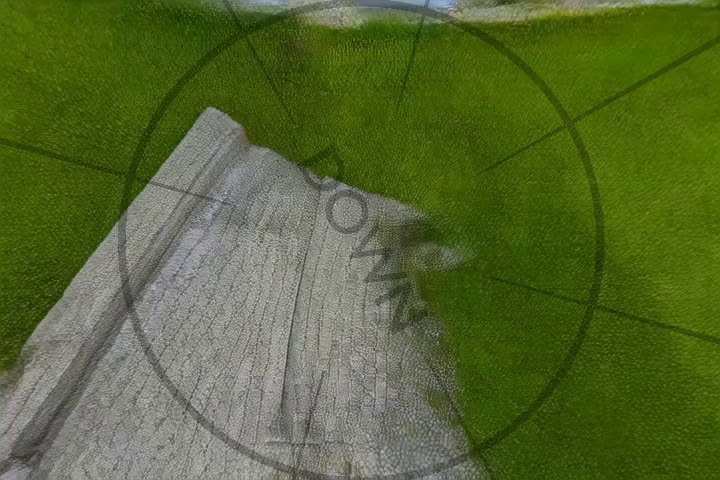} \\
    \multicolumn{4}{c}{\parbox{1.0\textwidth}{A minimalist bedroom corner features a white bedside table with a stool, a bed with brown and white linens, and a doorway in the background, bathed in soft, even light.}} &
    \multicolumn{4}{c}{\parbox{1.0\textwidth}{A peaceful park with a reflective pond, stone bridge, and lush trees, bathed in soft morning light, evokes calm and natural beauty.}} \\
    \hline
    \includegraphics[width=0.27\textwidth]{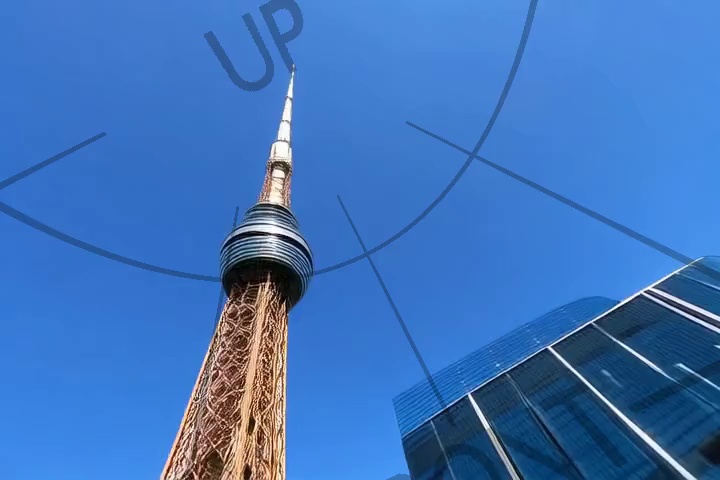} &
    \includegraphics[width=0.27\textwidth]{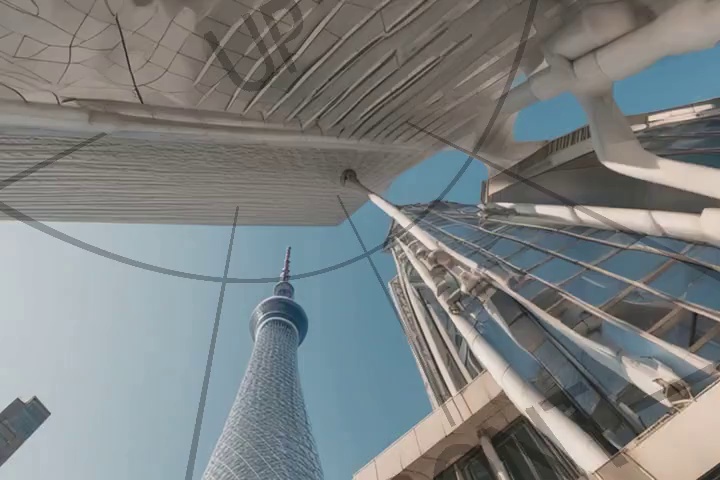} &
    \includegraphics[width=0.27\textwidth]{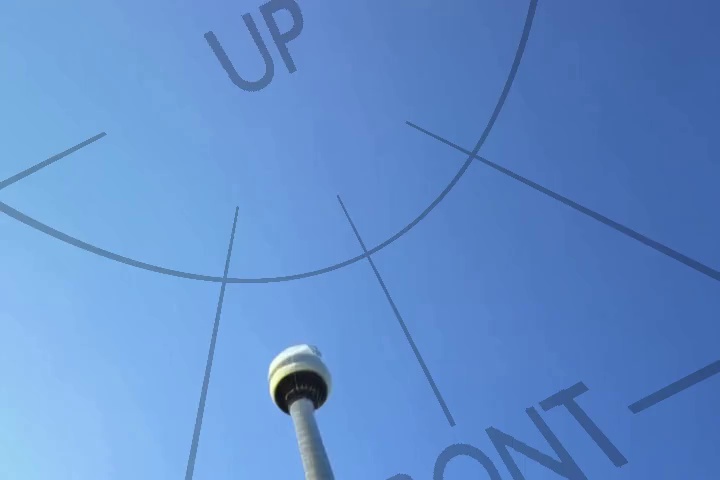} &
    \includegraphics[width=0.27\textwidth]{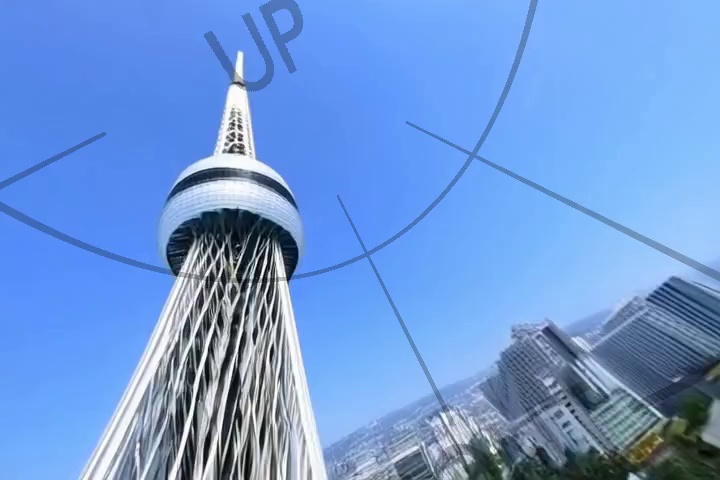} &
    \includegraphics[width=0.27\textwidth]{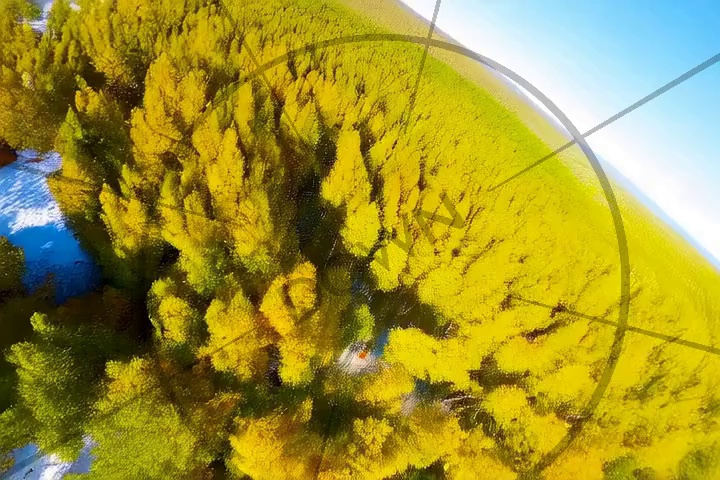} &
    \includegraphics[width=0.27\textwidth]{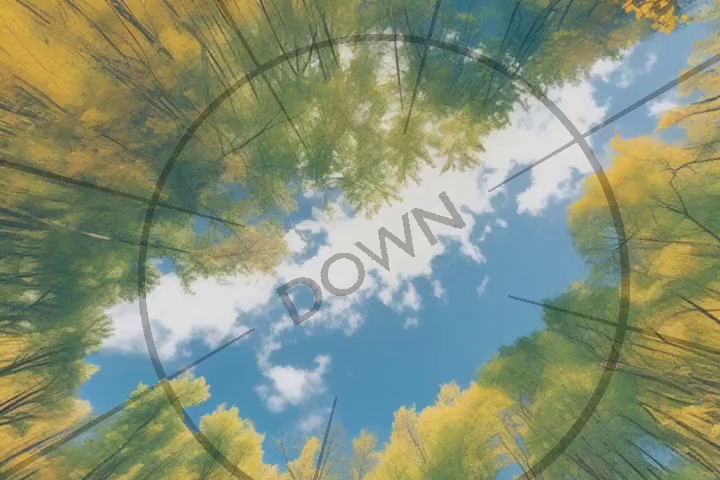} &
    \includegraphics[width=0.27\textwidth]{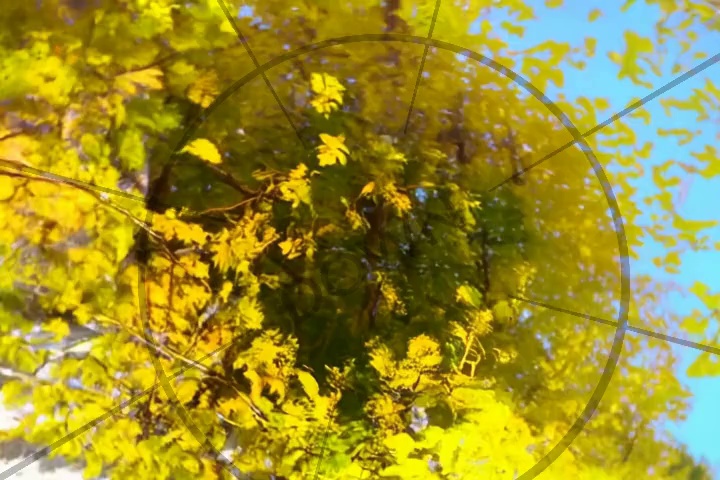} &
    \includegraphics[width=0.27\textwidth]{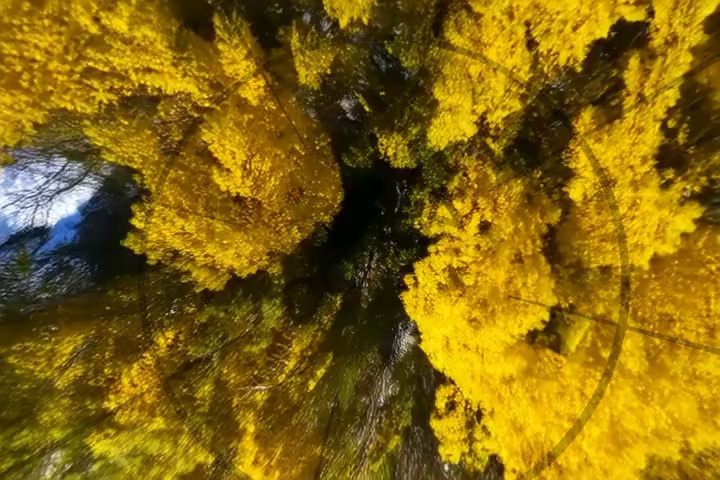} \\
    \includegraphics[width=0.27\textwidth]{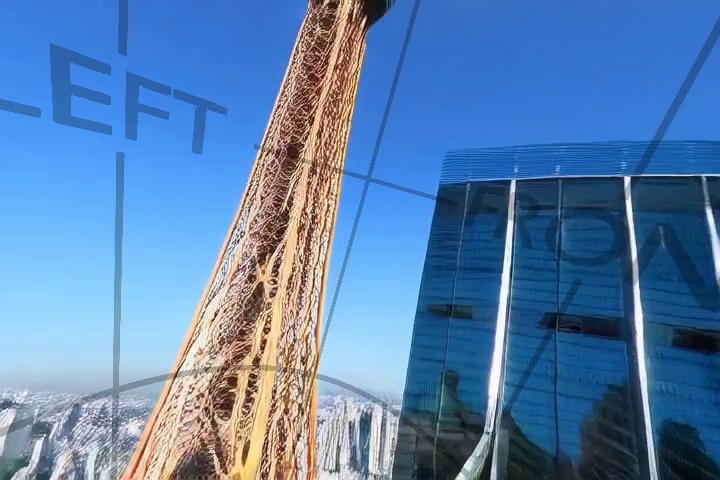} &
    \includegraphics[width=0.27\textwidth]{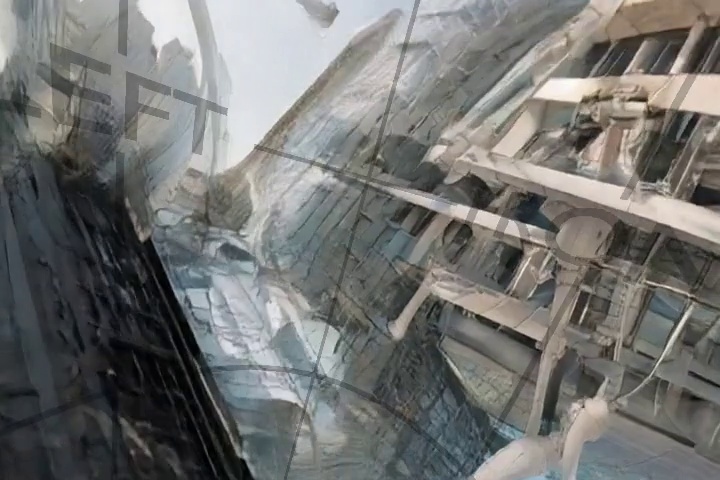} &
    \includegraphics[width=0.27\textwidth]{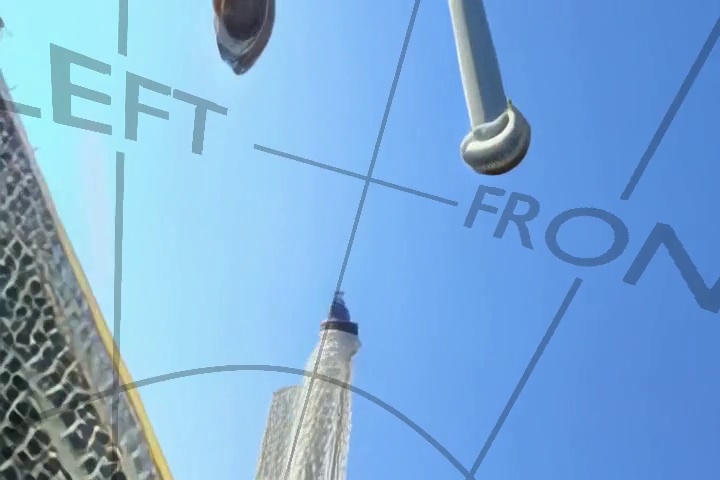} &
    \includegraphics[width=0.27\textwidth]{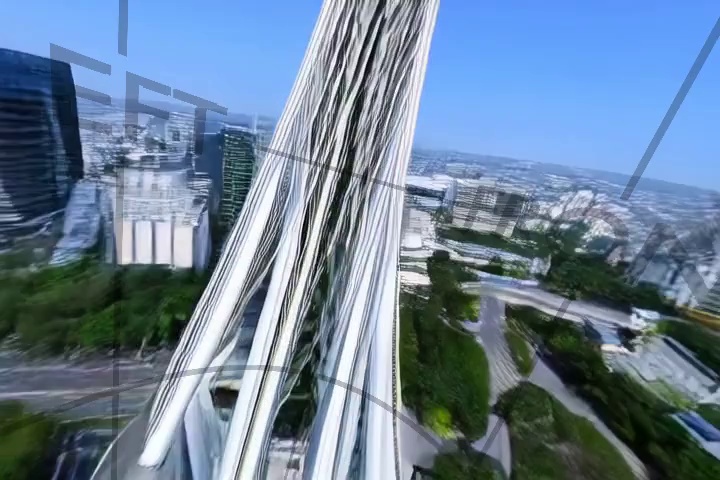} &
    \includegraphics[width=0.27\textwidth]{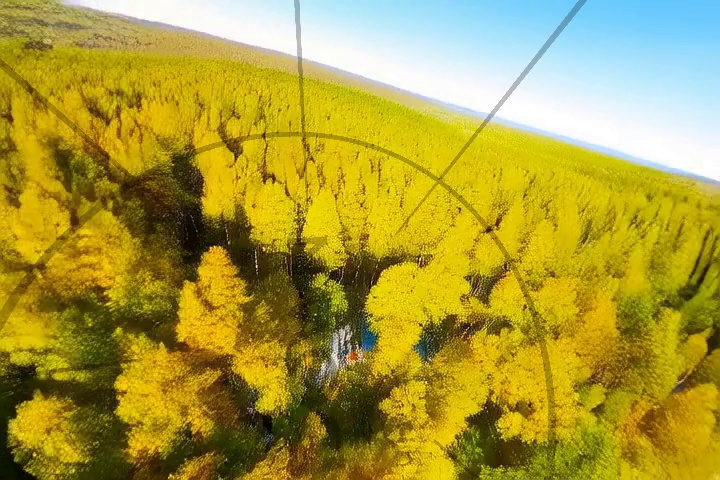} &
    \includegraphics[width=0.27\textwidth]{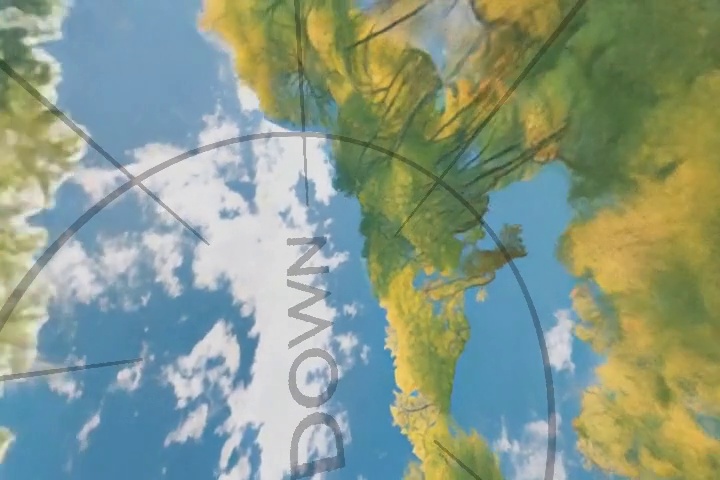} &
    \includegraphics[width=0.27\textwidth]{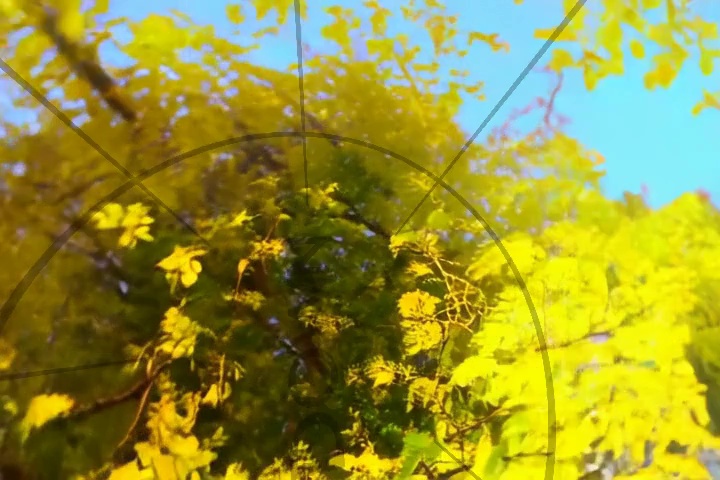} &
    \includegraphics[width=0.27\textwidth]{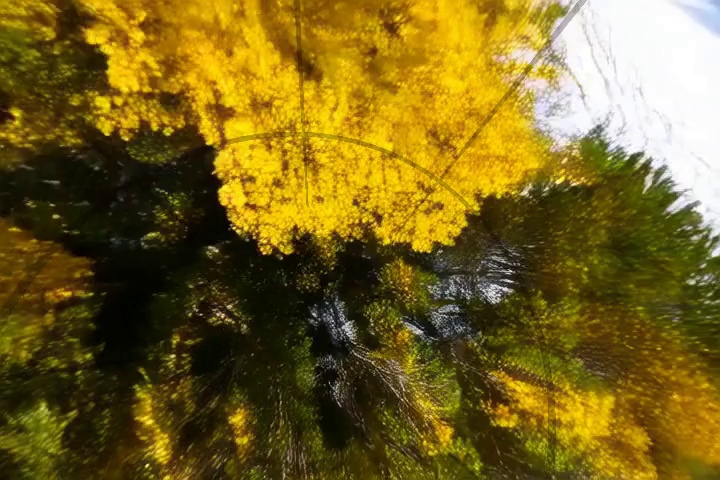} \\
    \includegraphics[width=0.27\textwidth]{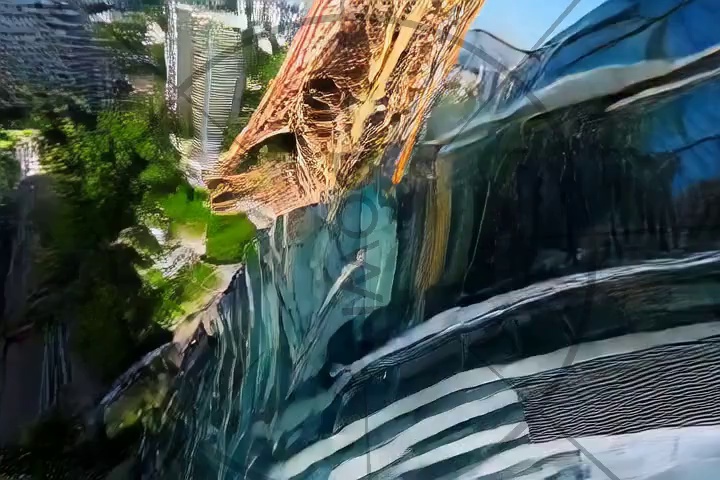} &
    \includegraphics[width=0.27\textwidth]{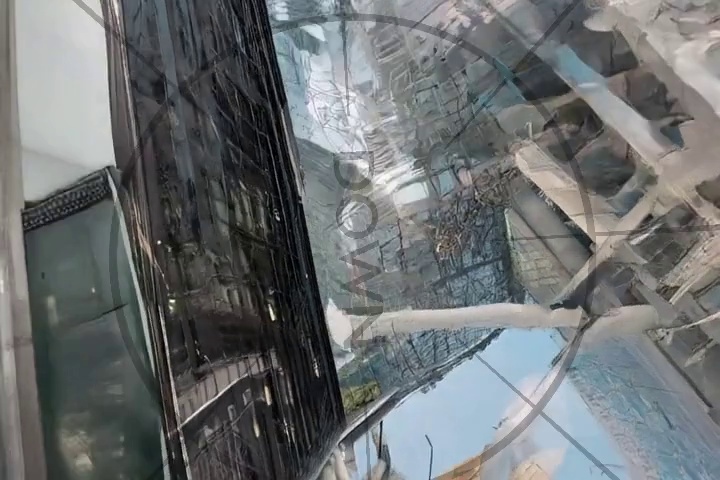} &
    \includegraphics[width=0.27\textwidth]{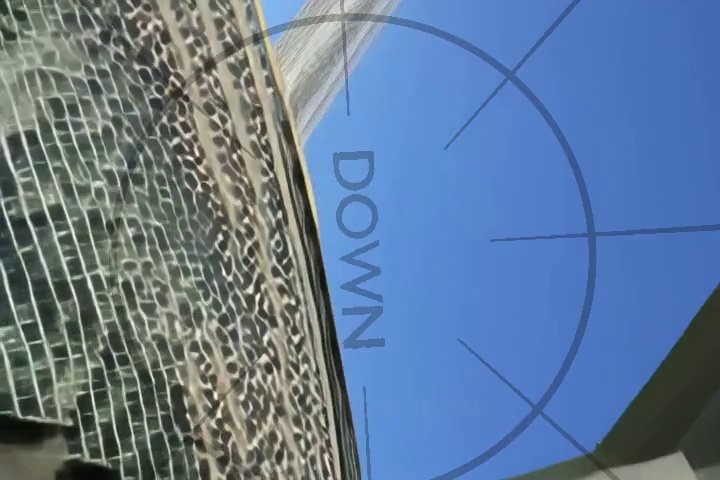} &
    \includegraphics[width=0.27\textwidth]{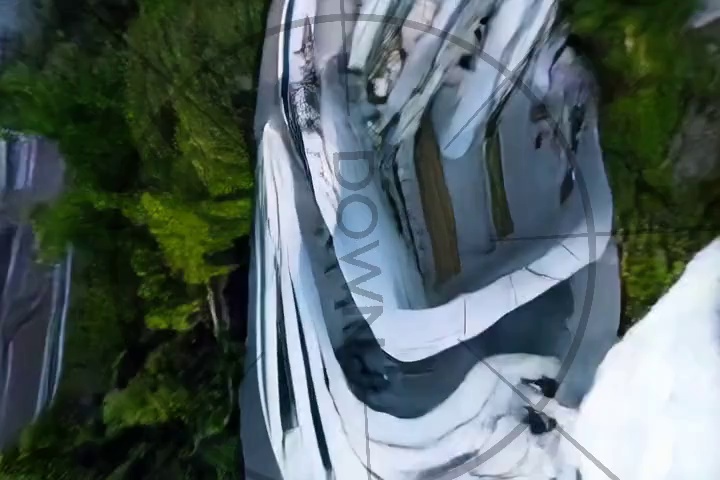} &
    \includegraphics[width=0.27\textwidth]{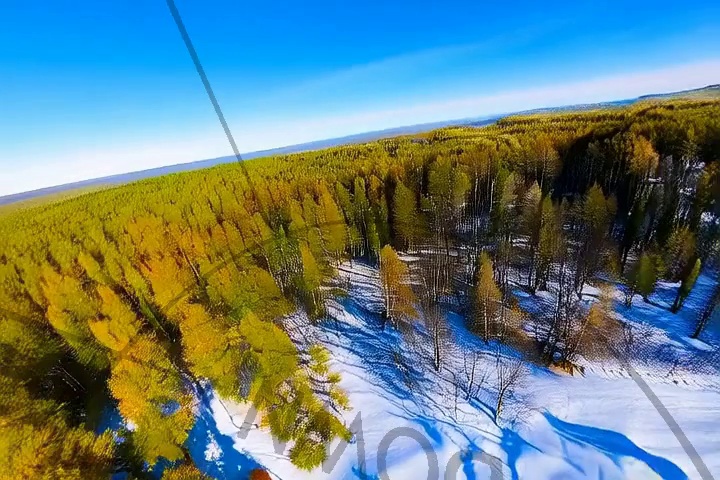} &
    \includegraphics[width=0.27\textwidth]{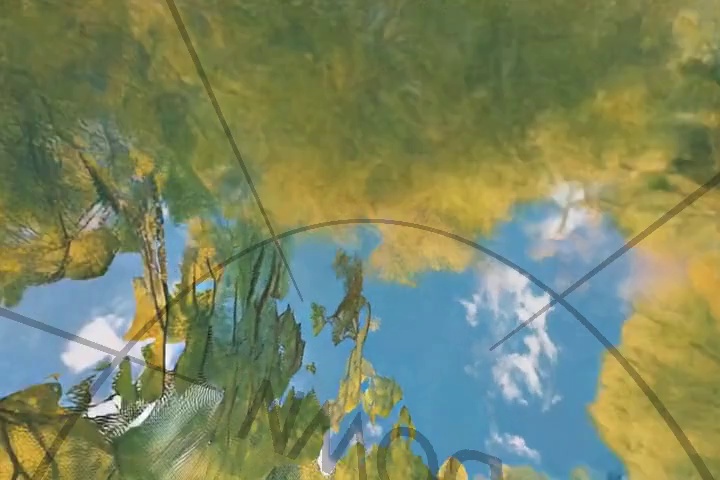} &
    \includegraphics[width=0.27\textwidth]{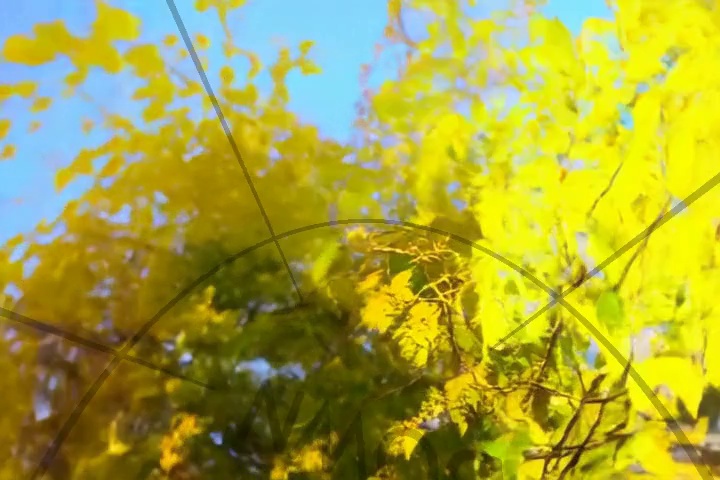} &
    \includegraphics[width=0.27\textwidth]{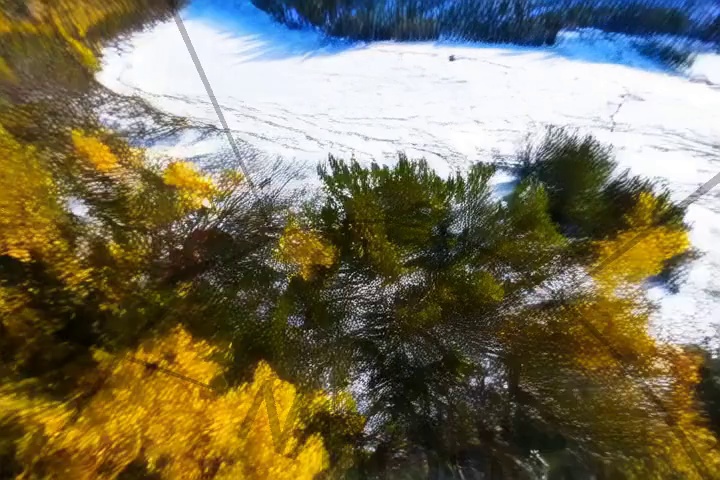} \\
    \multicolumn{4}{c}{\parbox{1.0\textwidth}{A towering Tokyo Skytree rises against a clear blue sky, flanked by a sleek modern building, evoking a sense of architectural wonder and urban beauty.}} &
    \multicolumn{4}{c}{\parbox{1.0\textwidth}{A vibrant forest view reveals yellow and green foliage and a bright blue sky, evoking tranquility and natural beauty before transitioning to a snowy woodland.}} \\
    \end{tabular}%
   }
    \caption{Qualitative results on the SpatialVID-extreme benchmark. The input absolute camera angle is shown as a dark overlay on the image. We observe that our method aligns well with the overlay, showing the sky when the zenith (``UP'') is in the view and showing the horizon aligned with the equator line (going through ``FRONT''), for example. Please refer to the supp.\ for additional results.}
    \label{fig:qualitative}
\end{figure*}

\myparagraph{Ablations.}

We trained separate \method checkpoints in which we ablated null-pitch conditioning (\ie, captioning is done on the training video itself) and absolute camera conditioning (\ie, making the camera poses relative to the first frame). We report the results in \cref{tab:quant_results}, in the ``w/o null-pitch cond.'' and ``w/o absolute rotations'' rows, respectively.
Making cameras relative to the first frame leads to a 3.67$\times$ and 3.36$\times$ increase in pitch and gravity error respectively.
This high error is due to the fact that the absolute extreme pitch and roll information are simply not provided to the model. 
Ablating the null-pitch conditioning leads to a 37\% and 72\% increase in pitch error, for the ``WAN 2.1 1.3B + UCPE encoding'' and ``WAN 2.2 5B + Plücker camera encoder'' variants, respectively. A similar effect is seen in the gravity error. We further investigate the effect of null-pitch conditioning in the next section.
\subsection{Prompt-camera entanglement evaluation}
\label{sec:prompt-camera-eval}

To investigate the effect of null-pitch conditioning, we develop a second benchmark dataset to evaluate whether methods can disambiguate between two possibly conflicting inputs: the prompt and the camera pitch angle. Indeed, pitch is highly correlated with semantic concepts, \eg, a +90\degree{} pitch (looking up) is strongly associated with images of the sky/ceiling, whereas a -90\degree{} pitch is strongly associated with ground features. Here, roll and yaw angles are ignored, since both are uncorrelated to object semantics in the field of view. 

We start by randomly selecting 20 panoramas from the PolyHaven dataset~\cite{polyhavenHDRIs2025}. For each, we take a forward-looking crop with a 90\degree{} FoV and caption it using InternVL-3-8B~\cite{zhu2025internvl3}. We then give this prompt to each evaluated method along with a static camera orientation. The methods are evaluated on pitch angles going from -90\degree{} to +90\degree{}, in 10\degree{} increments. In total, this benchmark consists of 380 pairs of input prompts and pitch angles. 

\begin{figure*}
\begin{minipage}{0.65\textwidth}
\centering

\tiny
\setlength{\tabcolsep}{1pt}
\begin{tabular}{
m{0.06\linewidth}
>{\centering\arraybackslash}m{0.175\linewidth}
>{\centering\arraybackslash}m{0.175\linewidth}
>{\centering\arraybackslash}m{0.175\linewidth}
>{\centering\arraybackslash}m{0.175\linewidth}
>{\centering\arraybackslash}m{0.175\linewidth}
}

 & PreciseCam & UCPE & w/o null-pitch & Ours & Ground truth \\

\rotatebox{90}{\shortstack{Up\\(+90\degree)}}
& \includegraphics[width=\linewidth]{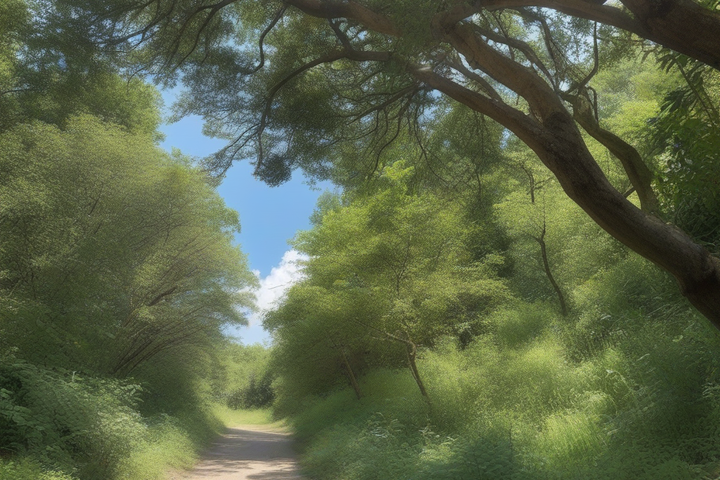}
& \includegraphics[width=\linewidth]{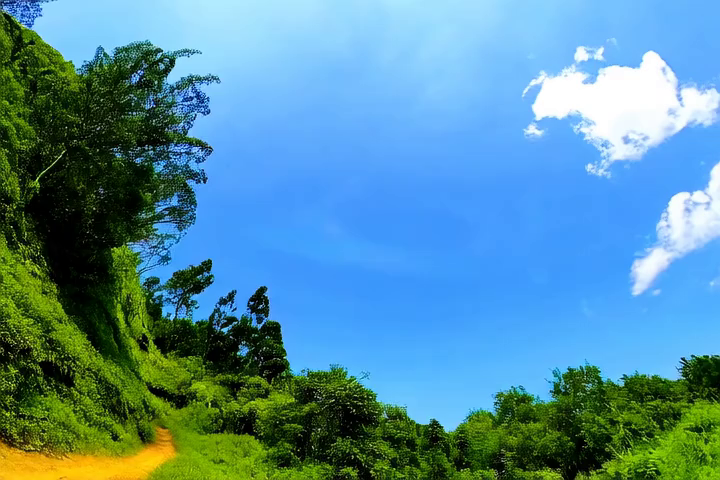}
& \includegraphics[width=\linewidth]{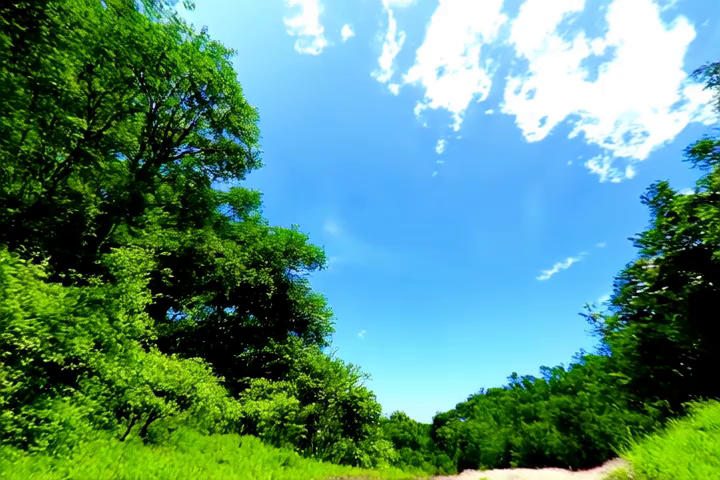}
& \includegraphics[width=\linewidth]{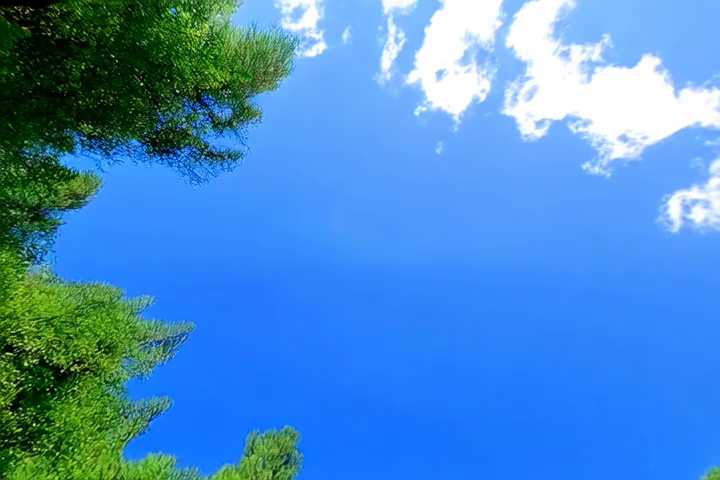}
& \includegraphics[width=\linewidth]{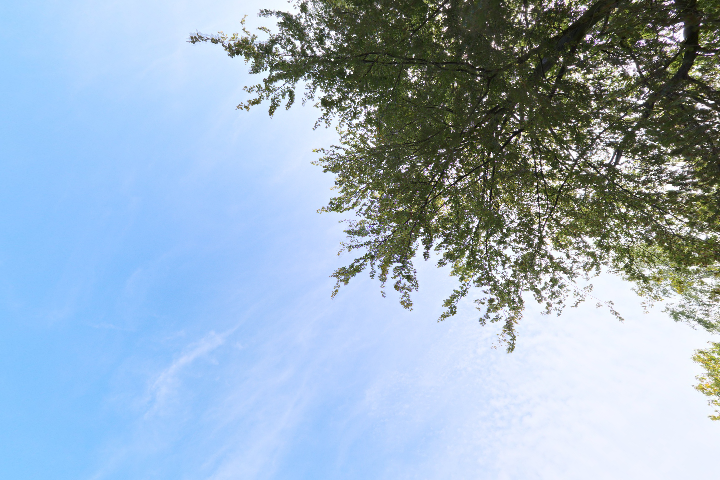}
\\

\rotatebox{90}{\shortstack{Forward\\(0\degree)}}
& \includegraphics[width=\linewidth]{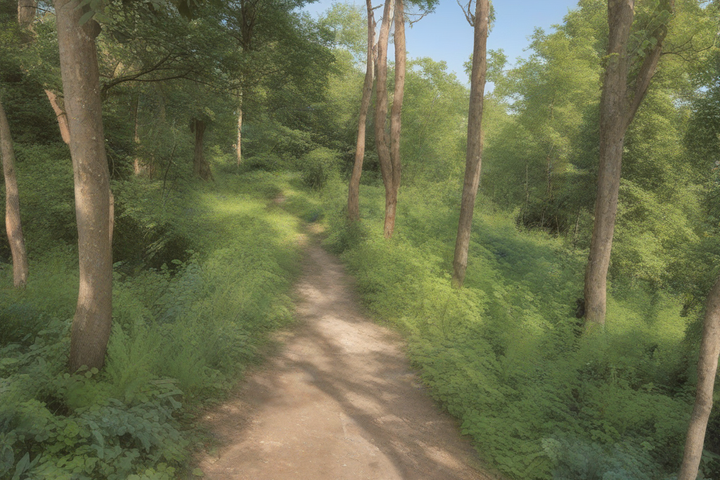}
& \includegraphics[width=\linewidth]{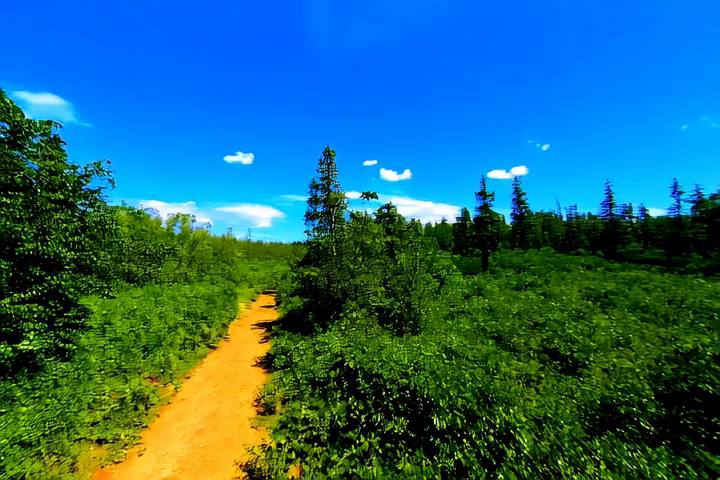}
& \includegraphics[width=\linewidth]{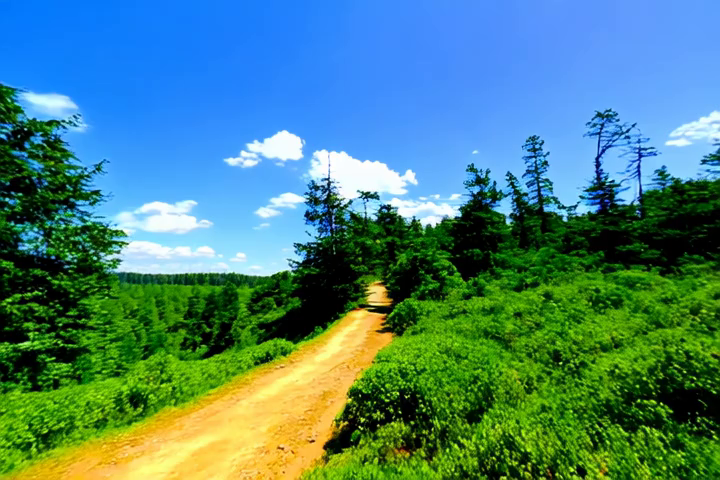}
& \includegraphics[width=\linewidth]{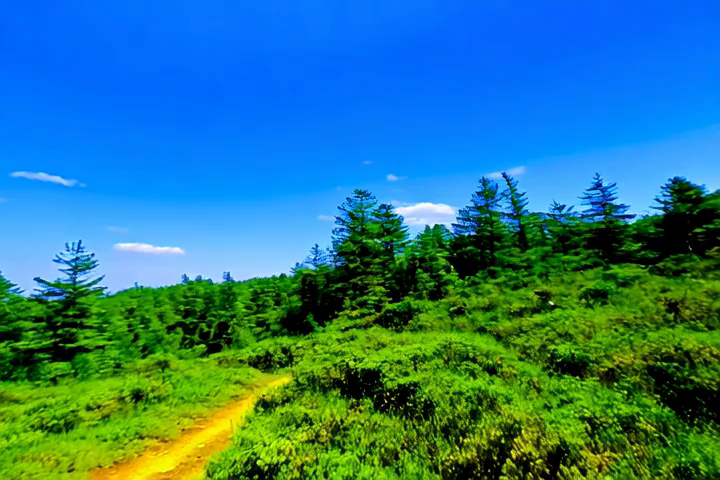}
& \includegraphics[width=\linewidth]{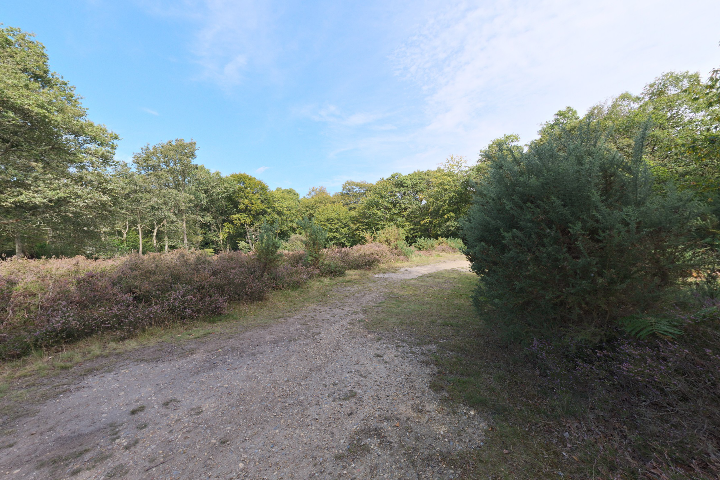}
\\

\rotatebox{90}{\shortstack{Down\\(-90\degree)}}
& \includegraphics[width=\linewidth]{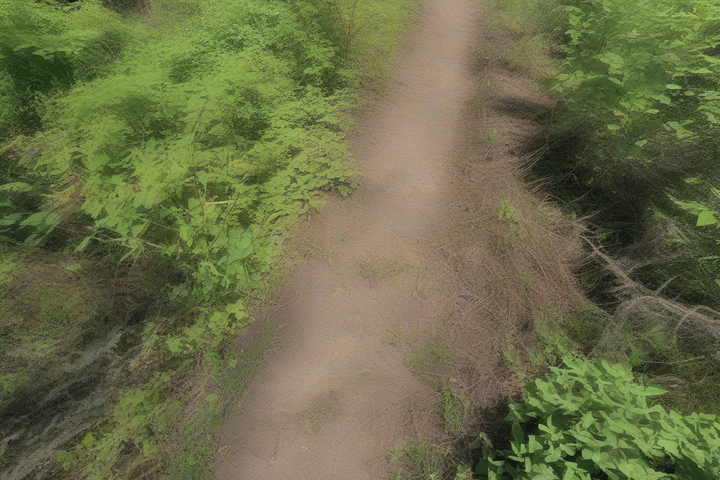}
& \includegraphics[width=\linewidth]{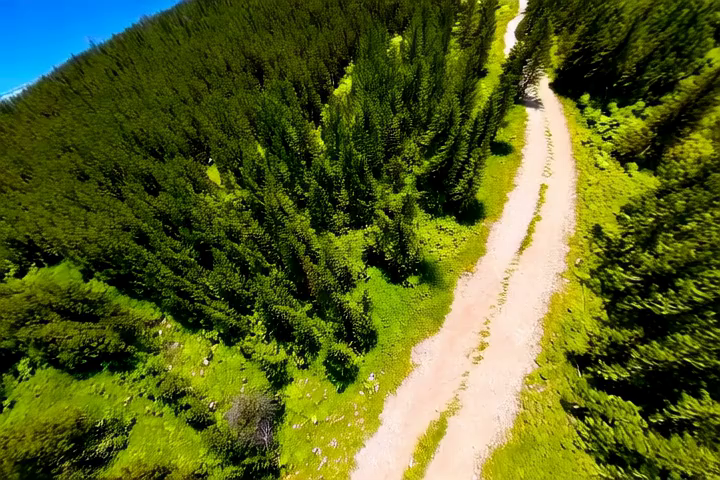}
& \includegraphics[width=\linewidth]{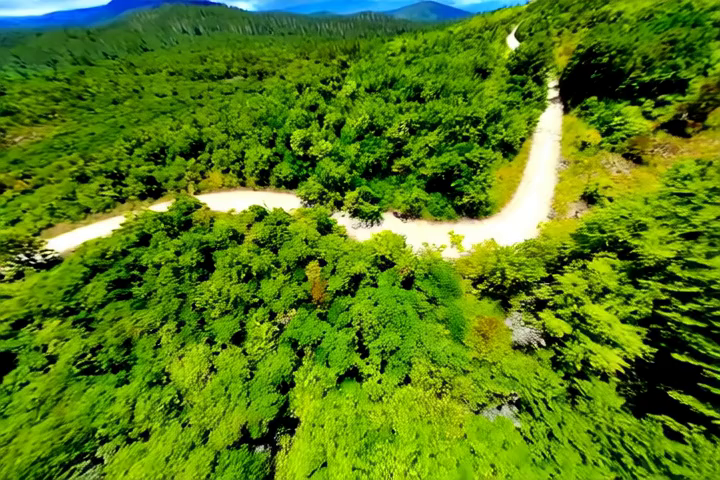}
& \includegraphics[width=\linewidth]{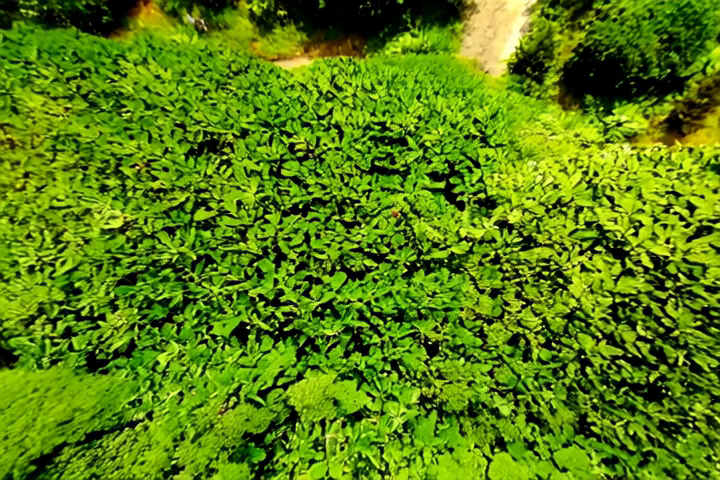}
& \includegraphics[width=\linewidth]{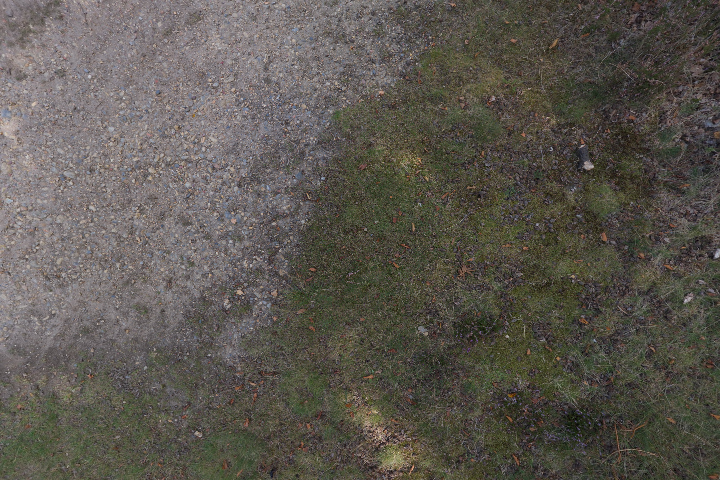}
\\
& \multicolumn{5}{p{0.88\textwidth}}{\centering
``A dirt path winds through a lush, green forest with tall trees and dense shrubs. The sky is clear and blue, with a few scattered clouds. The path leads into the distance, surrounded by vibrant vegetation.''
}
\end{tabular} \\
(a) Qualitative example from our entanglement benchmark
\end{minipage}
\hfill
\begin{minipage}{0.30\textwidth}
\centering
\tiny
\includegraphics[width=\linewidth]{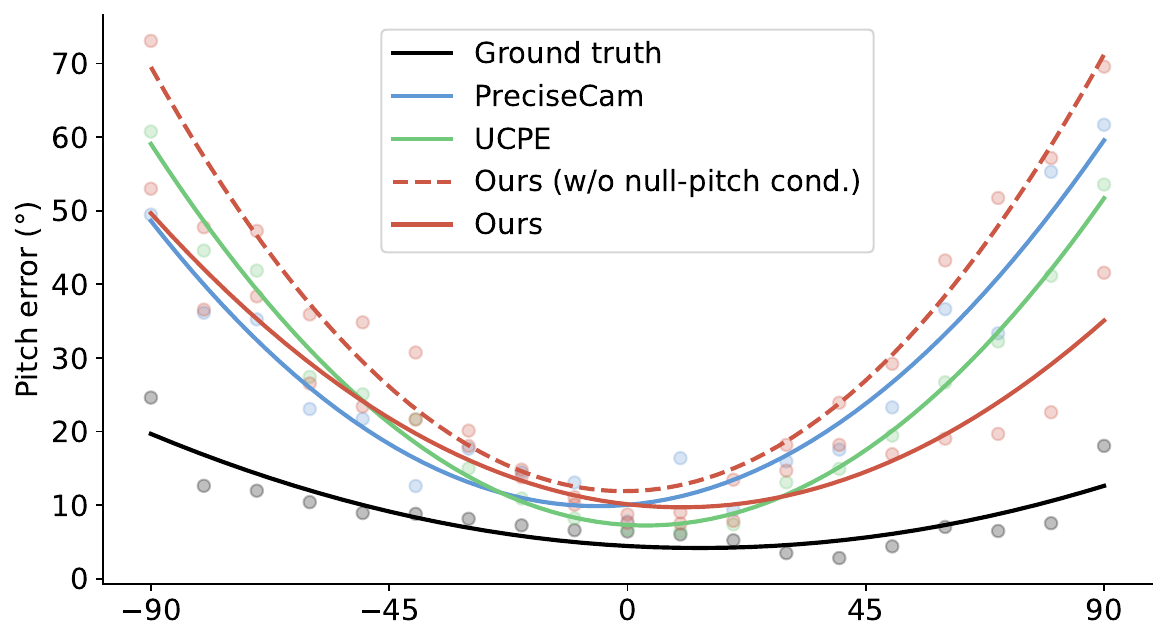}\\
\includegraphics[width=\linewidth]{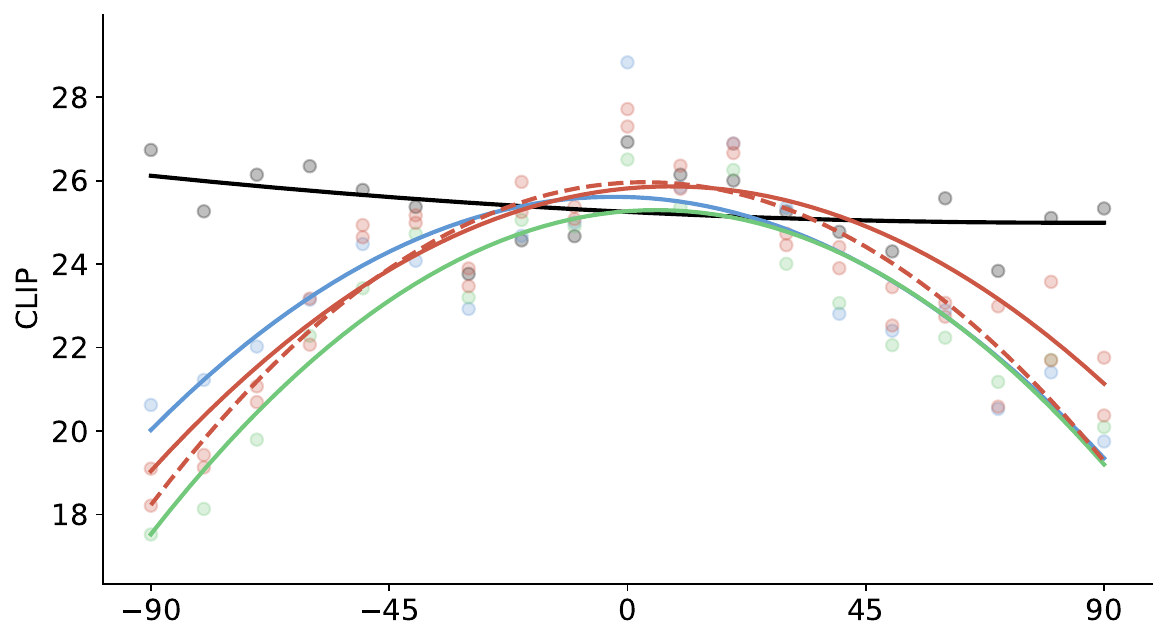}\\
(b) Pitch error and CLIP variation over our entanglement benchmark.
\end{minipage}

\caption{Evaluating prompt-camera pitch entanglement. (a) Without careful captioning, generative models may ignore camera conditioning when it conflicts with prompt semantics, especially at extreme pitch values (\eg,~-90\degree{} and +90\degree{}). (b) Null-pitch conditioning reduces pitch error at extreme angles (top) and improves CLIP similarity to the ground truth (bottom). See the supp. for additional results.}
\label{fig:entanglement_prompt_pitch}
\end{figure*}

The results obtained with baselines providing gravity-aligned camera control (UCPE~\cite{zhang2025unifiedcamerapositionalencoding} and PreciseCam~\cite{bernal2025precisecam}) are shown in \cref{fig:entanglement_prompt_pitch}. Qualitatively, we observe in \cref{fig:entanglement_prompt_pitch}-(a) that our proposed null-pitch conditioning enforces strict adherence to the camera orientation when the camera is pointing straight up or straight down. Quantitatively, we observe in \cref{fig:entanglement_prompt_pitch}-(b) (top) a substantially lower pitch-angle error across all 20 scenes, particularly at extreme absolute pitch values.

One simple way to assess the visual alignment between the text prompt and the output video is to compute the CLIP score directly between them. However, this naive approach has a key limitation: the prompt features may rightly be absent from the video if they violate the user-specified camera constraint (\eg, textual descriptions of ground content should not be visible if the camera is pointing up)---but not doing so would negatively affect the CLIP score. To address this contradiction, we compute the CLIP score between the generated video frames and a caption of the ground-truth  frame (obtained with InternVL-3-8B) for each pitch angle separately, and report results in \cref{fig:entanglement_prompt_pitch}-(b) (bottom). We observe that null-pitch conditioning \emph{improves} CLIP similarity at extreme pitch values, aligning results closer to the ground-truth curve. See the supplementary material for a detailed analysis.

\section{Discussion}

We present \method{}, a method that enables \emph{extreme and arbitrary camera control} in text-to-video generation by grounding camera poses in a gravity-aligned absolute coordinate system. Our pipeline combines 360\degree{} panoramic videos with camera pose estimation to produce training data covering the full sphere of possible viewpoints, including extreme pitch, large roll, and full 360\degree{} rotation trajectories that were previously beyond the reach of generative video models. We also proposed null-pitch conditioning, a data generation step that enhances camera conditioning fidelity when the text prompts contain conflicting cues. 

While it introduces novel capabilities for text-to-video models, our method still has a few limitations. First, our current on-the-fly camera sampling is restricted to rotations and must use the translations from the original source videos. A promising area for future research is to extend this data augmentation to include translations, potentially by leveraging recent successes in real-time novel view synthesis methods such as 3DGS~\cite{kerbl20233d}. Second, the generated videos can exhibit visual artifacts (for example, the distortions in the Tokyo Skytree in \cref{fig:qualitative}). We expect that continued progress in video generation models will mitigate these issues while remaining compatible with our framework. We hope our work lays the foundation for robust camera control under extreme rotations in video generation, empowering creators to produce dynamic storyboards with arbitrary camera orientations and convey their visual ideas with greater precision and speed.

\paragraph*{Acknowledgements.}
This research was supported by Adobe and a Natural Sciences and Engineering Research Council of Canada (NSERC) scholarship, ref. 600578. Computing resources were provided by Adobe and the Digital Research Alliance of Canada. We thank Yohan Poirier-Ginter, Qitao Zhao, Rahul Sajnani, Jack Hilliard and Jonathan Roussel for discussions and proofreading.

\bibliographystyle{splncs04}
\bibliography{main}
\end{document}